\documentclass[letterpaper]{article} 
\usepackage{aaai2026}  
\usepackage{times}  
\usepackage{helvet}  
\usepackage{courier}  
\usepackage[hyphens]{url}  
\usepackage{graphicx} 
\usepackage{natbib}  
\usepackage{caption} 
\usepackage{algorithm}
\usepackage{algorithmic}
\usepackage{graphicx}
\usepackage{natbib}
\usepackage{amsmath}
\usepackage{amssymb}
\usepackage{multirow}
\usepackage{booktabs}
\usepackage{subcaption}
\DeclareMathOperator*{\argmax}{argmax}
\usepackage{dsfont}

\usepackage{newfloat}
\usepackage{listings}
\DeclareCaptionStyle{ruled}{labelfont=normalfont,labelsep=colon,strut=off} 
\floatstyle{ruled}
\newfloat{listing}{tb}{lst}{}
\floatname{listing}{Listing}
\title{LLM Collaboration with Multi-Agent Reinforcement Learning}
\author{
    Shuo Liu,
    Tianle Chen\textsuperscript{\rm *},
    Zeyu Liang\textsuperscript{\rm *},
    Xueguang Lyu,
    Christopher Amato\textsuperscript{\rm $\dagger$}
}
\affiliations{
    Khoury College of Computer Sciences,\\
    Northeastern University\\
    Boston, MA, 02115, USA\\
    \{liu.shuo2, chen.tianle, liang.zey, lu.xue, c.amato\}@northeastern.edu
}

\usepackage{bibentry}

\begin{document}

\maketitle

\begin{abstract}
A large amount of work has been done in Multi-Agent Systems (MAS) for modeling and solving problems with multiple interacting agents. However, most LLMs are pretrained independently and not specifically optimized for coordination. For example, existing LLM fine-tuning frameworks rely on individual rewards, which require complex reward designs for each agent to encourage collaboration. To address this challenge, we model LLM collaboration as a cooperative Multi-Agent Reinforcement Learning (MARL) problem. We develop a multi-agent, multi-turn algorithm, Multi-Agent Group Relative Policy Optimization (MAGRPO), to solve it, building on current RL approaches for LLMs as well as MARL techniques. Our experiments on LLM writing and coding collaboration demonstrate that fine-tuning multiple LLMs with MAGRPO enables agents to generate high-quality responses efficiently through effective cooperation. Our approach opens the door to using MARL methods for LLM collaboration and highlights the associated challenges.

Our code is available at \fontsize{7.8pt}{0pt}{\url{https://github.com/OpenMLRL/CoMLRL}}.
\end{abstract}

\section{Introduction}

State-of-the-art LLMs have demonstrated remarkable capabilities across diverse domains \cite{llama3, gpt4, gemini}. To adapt to specific applications or align with human preferences, fine-tuning has emerged as a critical training stage. Compared to supervised fine-tuning, Reinforcement Learning (RL) enables more generalizable learning for complex, multi-turn tasks through human-aligned reward design, making it an important technique for fine-tuning \cite{ouyang, deepseek-r1, zieglerrlhf}.

Likewise, Multi-Agent Systems (MAS) have been extensively studied over the past decades, with substantial progress in modeling and solving problems involving multiple agents \cite{weiss1999multiagent, van2008multi, yoavmas, peterstonemas}. In particular, advances in cooperative MAS have demonstrated strong potential for enabling effective collaboration in distributed settings, such as games, robotics, and traffic control \cite{smac, smacii, openai5, traffic, liutraffic, IJRR2025}. These developments motivate the application of MAS principles and techniques to LLM collaboration, where multiple LLMs working together can solve more complex tasks more robustly and efficiently.

There has been some recent work on coordinating multiple LLMs. Some approaches implement coordination at the inference stage, enabling agents to interact through debate, discussion, or verification \cite{dudebate, autogen, duverifier}. These methods operate at the prompt level, with fixed models that are not tuned toward coordination-centric objectives. The agents may have conflicting answers or spread incorrect information to other participants, limiting performance \cite{whymultillmfail, debatefail}. Moreover, the design of effective prompts remains difficult and unclear. Other approaches fine-tune agents independently with individual or role-conditioned rewards. However, they require carefully curated rewards for each individual or role \cite{spiral, maft}, and, as independent learning methods,  lack convergence guarantees \cite{Tan1993}.

We model LLM collaboration as a cooperative MARL problem \cite{marl-book,marl} and formalize it as a Decentralized Partially Observable Markov Decision Process (Dec-POMDP) \cite{decpomdp}. In LLM collaboration, multiple trainable LLMs generate responses synchronously based on their individual prompts. The external environment evolves according to the joint responses until the dialog ends. This general model allows a wide range of problems to be modeled and solved using versions of MARL algorithms. 
Following the efficient practice of Group Relative Policy Optimization (GRPO)~\cite{deepseek-math}, we propose Multi-Agent GRPO (MAGRPO) that trains LLMs in a multi-turn setting. MAGRPO leverages centralized group-relative advantages for joint optimization, while preserving decentralized execution for each agent. The resulting method builds off of state-of-the-art LLM approaches in GRPO and MARL approaches for centralized training and decentralized execution, such as MAPPO \cite{mappo}. Our experiments demonstrate that MAGRPO is able to learn different cooperation schemes while producing efficient and high-quality responses. 

Our contributions can be summarized as follows: (i) We model the LLM collaboration as a cooperative MARL problem, where multiple LLMs cooperate to generate joint responses; (ii) We develop the MAGRPO algorithm, which optimizes agent cooperation through aligned rewards while maintaining decentralized execution to maintain efficiency; (iii) Our experiments demonstrate that fine-tuning with MAGRPO improves both response efficiency and quality in writing and coding collaboration; (iv) We provide an analysis of the limitations of existing approaches and outline open challenges in applying MARL to LLM collaboration.

\section{Related Work}

\subsubsection{Test-Time Multi-Agent Interaction}
Recent work employs multiple agents with specialized roles interacting through diverse pipelines at test-time to enhance response quality. In multi-agent debate, agents iteratively formulate positions by reviewing other agents' outputs, where the final decision or answer is determined by majority voting or a summarizer \cite{dudebate, chandebate, liangdebate}. Role-based approaches allocate tasks across specialized agents \cite{autogen, chatdev, metagpt}. An agent may function as a verifier to assess the correctness of outputs \cite{skretaverifier, duverifier}, or as a macro-planner to orchestrate workers' responses. However, these multi-agent frameworks rely on prompt-level interactions among agents, often leading to ineffective communication and computational inefficiency. Moreover, the design of effective prompts and role assignment remains unclear, as prompts usually fail to reliably guide agent behavior, enforce role adherence, or support coherent coordination across tasks. These limitations motivate us to fine-tune LLMs in MAS to improve their cooperation.

\subsubsection{Multi-Agent Fine-Tuning}

Recent work has explored fine-tuning LLMs to improve their performance across diverse domains, e.g., arithmetic reasoning, navigation, and hidden-role games \cite{cory, amongus}. These approaches typically employ individual rewards or rewards conditioned on specific roles \cite{spiral, maft, marti2025}. Such reward structures often require careful manual specification, and their underlying rationale is rarely well justified. The misaligned or conflicting incentives can hinder effective coordination. Moreover, these methods lack convergence guarantees, as each agent learns independently in a non-stationary environment where other agents are simultaneously updating their policies. In this paper, we focus on cooperative scenarios, where LLMs are trained with verified, human-aligned rewards.

\section{Cooperative MARL for LLM Collaboration}

\begin{figure*}[t]
    \centering
    \includegraphics[width=0.94\textwidth]{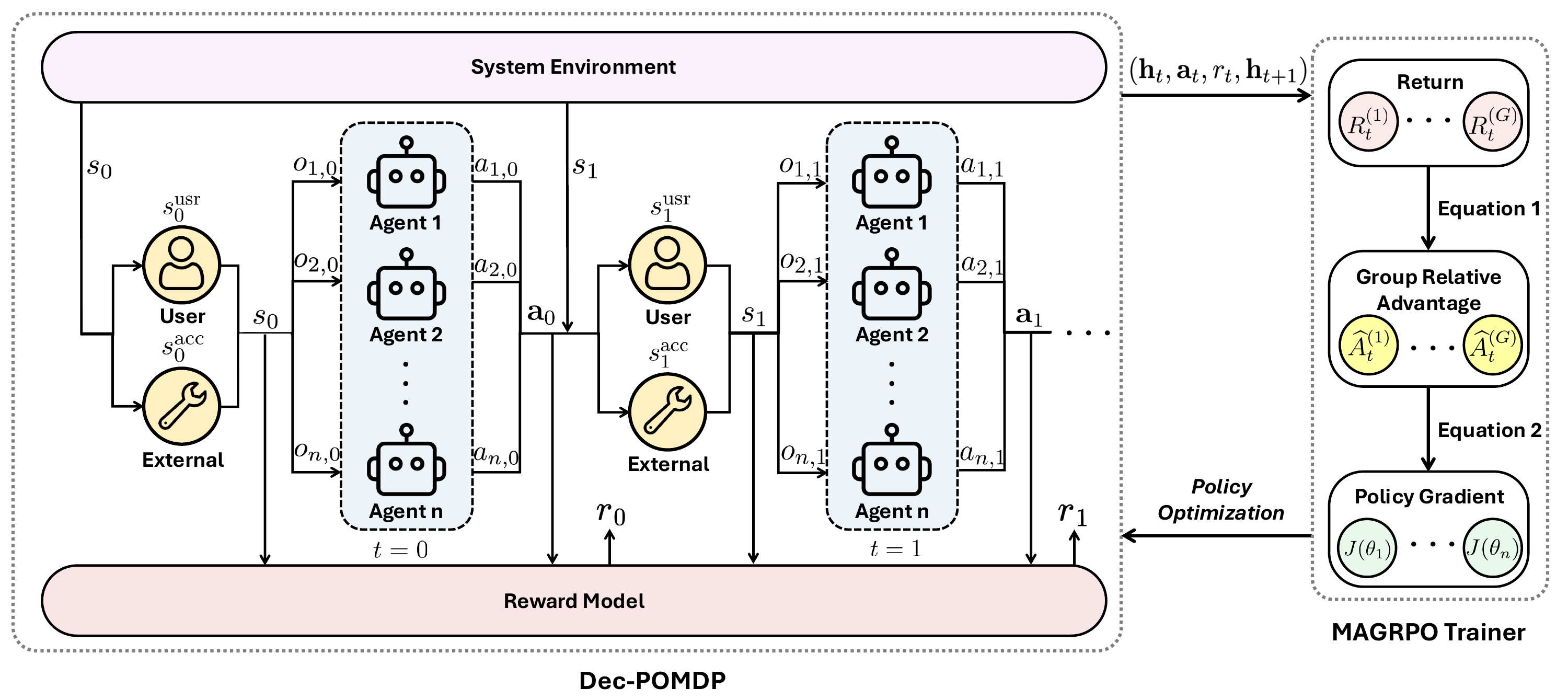}
    \caption{Illustration of Dec-POMDP and our MAGRPO algorithm.}
    \label{fig:mlrl}
\end{figure*}

Since LLMs can act as a special class of agents, we leverage advances in MAS to improve their collaboration. We model LLM collaboration as a cooperative MARL problem and outline its unique challenges. We formalize this problem as a Dec-POMDP, as shown in Figure~\ref{fig:mlrl}.

\subsection{LLM Collaboration}

LLM collaboration is the problem where LLMs cooperatively solve a class of tasks in MAS. Tasks are specified in language and provided to the LLMs as prompts. Each LLM generates a response synchronously in response to its own instructions. All responses jointly form a solution to the task.

Most tasks cannot be resolved in one turn. Users, external models, or systems validate the solutions and provide additional requirements or suggestions for LLMs. These components also serve as part of the environment for LLM collaboration, whose states may change based on the agents' outputs. The updates are embedded into prompts for subsequent turns. This iterative process continues until the task is successfully completed or a predefined turn limit is reached.

As discussed by a number of companies \cite{nvidia_llm_agents, anthropic_multi_agent}, a team of agents could generate a complex codebase. The code would be difficult, costly, and time-consuming to generate with a single agent, but a group of LLMs could do so quickly and cheaply. None of these agents is self-interested, but they can be trained using a scheme such as the one in this paper. Using a joint reward allows agents to specialize as needed to complete the task without complex prompt or reward engineering.

\subsection{Problem Formalization}

We formalize LLM collaboration as a subclass of the cooperative MARL problem, considering LLMs are agents and the types of problems they are solving. This problem is a form of Dec-POMDP \cite{decpomdp}, which allows cooperation through a joint reward while preserving scalable decentralized control. We show 2 instantiations of our framework in writing and coding tasks in the experiments.

Mathematically, our LLM Dec-POMDP is defined by a tuple $\langle \mathcal{I}, \mathcal{S}, \{\mathcal{O}_i\}, \{\mathcal{A}_i\}, R, T, H \rangle$.
\begin{itemize}
    \item $\mathcal{I} = \{1,\cdots, n\}$ denotes the set of $n$ LLM agents, each instantiated with a pre-trained language model.
    \item $\mathcal{S}$ denotes the full global state space. At turn $t$, a full state $s_t= (s_t^{\text{acc}}, s_t^{\text{usr}})$ consists of parts that are accessible in the model and provided to the reward model $s_t^\text{acc} \in \mathcal{S}^\text{acc}$ (e.g., external models or systems), and the inaccessible user state $s_t^\text{usr} \in \mathcal{S}^\text{usr}$ that updates over time but is not maintainable. In a Dec-POMDP, the state can not be directly observed by the agents.
    \item $\mathcal{O}_i$ is the observation space for agent $i$ with $\mathcal{O}=\times_i \mathcal{O}_i$ the joint observation space. A local observation $o_{i,t}$ consists of natural language instructions (i.e., prompts), providing a partial and noisy view of $s_t$.
    \item $\mathcal{A}_i$ is the action space for agent $i$ with $\mathcal{A}=\times_i \mathcal{A}_i$ the joint action space. A local action $a_{i,t}$ is a response in natural language to the given prompt. 
    \item $R: \mathcal{S}^\text{acc} \times \mathcal{A} \rightarrow \mathbb{R}$ is the joint reward function implemented via predefined rules or a pretrained reward model. At turn $t$, the joint rewards $r_t$ are determined by the accessible part of current state $s^\text{acc}_t$ and the agents’ joint action $\mathbf{a}_t=\{a_{1, t}, \cdots, a_{n, t}\}$.
    \item $T: \mathcal{S} \times \mathcal{A} \rightarrow \Delta(\mathcal{S})$ is the underlying stochastic state transition function. At turn $t$, the agents’ joint actions $\mathbf{a}_t$ induce a shift to a new state $s_{t+1}\sim T(\cdot | s_t, \mathbf{a}_t)$, which reflects the updates in the user state and the states of external models and systems.
    \item $H$ is the episode horizon, i.e., the turn limit of the dialog.
\end{itemize}

In Dec-POMDP, since the states are not directly observed, each agent maintains its local observation-action history $\mathbf{h} = \{h_1, \cdots, h_n\}$ to infer information about the state. A solution to a Dec-POMDP is a joint policy that maximizes the expected cumulative reward, $\boldsymbol{\pi}^* = \{\pi_1^*, \cdots , \pi_n^*\}=\arg\max_{\boldsymbol{\pi}} \mathbb{E}_{\boldsymbol{\pi}}\left[\sum_{t=0}^{H-1} R(s^\text{acc}_t, \mathbf{a}_t)\right]$. A joint policy is a set of local policies $\pi_i$, which condition on the local observation-action history $h_{i,t} = \{o_{i,0}, a_{i,0}, \ldots, o_{i,t}\}$.

RL methods for Dec-POMDPs have become a popular topic (e.g., \cite{COMA, MADDPG, Mordatch18, QMIX, QPLEX, mappo, marl-book, lyu2023centralized, Marchesini25}) with methods successful at scaling to large state, action, and observation spaces.
Many methods use Centralized Training for Decentralized Execution (CTDE), where they use some centralized information during training (e.g., a centralized value function estimate) but are still able to execute in a decentralized manner when training is completed \cite{CTDE}.  

\subsection{Challenges in LLM Collaboration}

LLM collaboration presents unique challenges compared to traditional MARL problems, where LLM agents receive and process tasks through natural language.

\subsubsection{Representations in Natural Language}

Unlike traditional cooperative MARL agents, LLM agents operate over natural language, receiving instructions and generating responses as sequences of tokens. MARL approaches could model this problem at the token or sequence level. At the token level, the number of actions and observations is smaller, but the problem horizon can be very long. At the sequence level, the action and observation space is much larger, but the horizon is much shorter. Moreover, token-level rewards are often uninformative, as both queries and responses must form coherent and semantically meaningful structures. As adopted in prior RL methods \cite{ouyang, dpo}, we model each agent's decision-making process as a direct mapping from input instructions to complete responses to enable efficient and stable training. Nevertheless, the best modeling and solution approaches remain an open question. 

\subsubsection{Training Paradigm}
As mentioned above, many MARL methods use centralized training for decentralized execution (CTDE). Unfortunately, standard CTDE methods use centralized value models in the form of centralized critics \cite{COMA, MADDPG, mappo} or mixers in value decomposition methods \cite{QMIX, QPLEX}, which allow additional information during training but do not scale well to very large action and observation spaces (such as those in LLM collaboration). Conversely, Decentralized Training and Execution (DTE) methods \cite{marl} train a set of models, one for each agent in a decentralized manner, which are typically more scalable but do not use additional information during training (even when it is available). 
It is an open question which paradigm to use to maximize performance while maintaining scalability in the LLM collaboration problem. 
In this paper, we balance decentralized execution with centralized training using group-based Monte Carlo estimates. Experiments show the effectiveness of our approach on short-horizon tasks. 

\section{MAGRPO}

We propose the Multi-Agent GRPO (MAGRPO) algorithm to jointly train LLM agents in MAS while maintaining decentralized execution.

\begin{algorithm}[t]
\caption{MAGRPO}
\begin{algorithmic}[1]
\REQUIRE Dataset $\mathcal{D}$, $n$ pretrained LLMs with policies $\{\pi_{\theta_1}, \cdots, \pi_{\theta_n}\}$, reward model $R$, generation group size $G$, learning rate $\alpha$
\FOR{each episode}
\STATE Sample a task $\sim \mathcal{D}$
\STATE Initialize observations $o_{i, 0}, \forall i\in \mathcal{I}$, according to the task, and $\mathbf{o}_0=\{o_{1, 0}, \cdots, o_{n, 0}\}$
\STATE $h^{\mathcal{G}}_{i, 0}\gets o_{i, 0}$, $\forall i \in \mathcal{I}$, and $\mathbf{h}^{\mathcal{G}}_0=\{h^{\mathcal{G}}_{1, 0}, \cdots, h^{\mathcal{G}}_{n, 0}\}$
	\FOR{turn $t=0$ to $H-1$}
        \STATE Generate a group of responses $a^{\mathcal{G}}_{i, t} \gets \pi_{\theta_i}(\cdot | h^{\mathcal{G}}_{i, t})$, $\forall i \in \mathcal{I}$, where $h^{\mathcal{G}}_{i, t}=\{h^{(1)}_{i, t}, \cdots, h^{(G)}_{i, t}\}$, $a^{\mathcal{G}}_{i, t}=\{a^{(1)}_{i, t}, \cdots, a^{(G)}_{i, t}\}$, and $\mathbf{a}^{\mathcal{G}}_0=\{a^{\mathcal{G}}_{1, t}, \cdots, a^{\mathcal{G}}_{n, t}\}$
    \STATE Obtain joint rewards $r^{\mathcal{G}}_{t}$ from system
      \STATE Receive new observations $o^{\mathcal{G}}_{i, t+1}$, and update history $h^\mathcal{G}_{i, t+1} \gets \{h^\mathcal{G}_{i, t}, a^\mathcal{G}_{i, t}, o^\mathcal{G}_{i, t+1}\}$, $\forall i \in \mathcal{I}$
    \ENDFOR
    \FOR{turn $t=H-1$ to $0$}
    \STATE Calculate return $R^{(g)}_t\gets \sum_{\tau=t}^{H-1} r^{(g)}_\tau$, $\forall g \in \mathcal{G}$
    \STATE Estimate $\widehat{A}^{(g)}_{t}$, $\forall g \in \mathcal{G}$ according to Equation~\ref{eq:advantage}
        	\STATE Calculate $J(\theta_i)$, $\forall i \in \mathcal{I}$ according to Equation~\ref{eq:calculate_gradient}
        	\STATE $\theta_i \gets \theta_i + \alpha \nabla_{\theta_i} J(\theta_i)$, $\forall i \in \mathcal{I}$
    \ENDFOR
\ENDFOR
\RETURN $\boldsymbol{\pi}_{\boldsymbol{\theta}}=\{\pi_{\theta_1}, \cdots, \pi_{\theta_n}\}$
\end{algorithmic}
\label{alg:multi-turn}
\end{algorithm}

Algorithm~\ref{alg:multi-turn} shows the procedure of MAGRPO. Given a dataset $\mathcal{D}$ containing task information (e.g., the descriptions of coding problems), $n$ LLMs that are optimized, each with a policy parameterized by $\theta_i$ and guided by a (shared) reward model $R$. In each episode, a task is sampled from the given dataset $\mathcal{D}$, which is used to construct initial observations $\mathbf{o}_0=\{o_{1, 0}, \cdots, o_{n, 0}\}$ and histories $\mathbf{h}_0=\{h_{1, 0}, \cdots, h_{n, 0}\}$. 
Taking inspiration from the single-agent GRPO algorithm \cite{deepseek-math}, 
at each turn $t$, each agent takes action by generating a group of responses $a^{\mathcal{G}}_{i, t}=\{a^{(1)}_{i, t}, \cdots, a^{(G)}_{i, t}\}$ following its policy $\pi_{i}(\cdot|h_{i, t}^\mathcal{G})$ based on its observation-action history $h_{i, t}^\mathcal{G}=\{h^{(1)}_{i, t}, \cdots, h^{(G)}_{i, t}\}$. 
The actions of individual agents are aggregated to form a group of joint actions $\mathbf{a}^{\mathcal{G}}_{t}=\{a^{\mathcal{G}}_{0, t}, \cdots, a^{\mathcal{G}}_{n, t}\}$. The agents receive a group of joint rewards $r^{\mathcal{G}}_{t}$ for their responses $\mathbf{a}^{\mathcal{G}}_{t}$, which conditions on the accessible part of the state $R(\cdot|s^{\text{acc}, \mathcal{G}}_t, \mathbf{a}^{\mathcal{G}}_{t})$. The joint actions triggers the transition $T(\cdot | s^{\mathcal{G}}_{t}, \mathbf{a}^{\mathcal{G}}_{t})$, where agents receive new observations $o^{\mathcal{G}}_{i, t+1}=\{o^{(1)}_{i, t+1}, \cdots, o^{(G)}_{i, t+1}\}$ and use them to construct histories $h^{\mathcal{G}}_{i, t+1}=\{h^{\mathcal{G}}_{i, t}, a^{\mathcal{G}}_{i, t}, o^{\mathcal{G}}_{i, t+1}\}$. This process continues until terminated at turn $H$ or the task is completed.

We employ stochastic gradient descent to train agents at the end of each episode. Without explicit value models, estimating history-action values incurs high variance. To stabilize training, we estimate the expected return of the current history by averaging over a group of Monte Carlo samples of the joint return $R^{\mathcal{G}}_{t}=\{R^{(1)}_{t}, \cdots, R^{(G)}_{t}\}$. As a result, we are able to generate a centralized estimate (common in MARL) without a large value model. For each turn $t$, the advantage of each joint action in the group is calculated as,
\begin{equation}
\widehat{A}^{(g)}_{t} = R^{(g)}_t -\frac{1}{G}\sum_{g=1}^{G}R^{(g)}_t,
\label{eq:advantage}
\end{equation}
where $R^{(g)}_t = \sum_{\tau=t}^{H-1} r^{(g)}_\tau$. Inspired by GRPO \cite{deepseek-math, drgrpo}, and MAPPO \cite{mappo}, the centralized advantage values can be used to update policy $\pi_i$ (parameterized by $\theta_i$) for each agent $i$. MAGRPO does not have importance sampling and thereby epsilon clipping for simplicity, and the KL divergence coefficient is set to be 0 to encourage greater policy deviation from the base model,
\begin{equation}
    J(\theta_i) =\mathbb{E}_{\mathbf{o}_0 \sim \mathcal{D}, \mathbf{h}^\mathcal{G} \sim \boldsymbol{\pi}_{\boldsymbol{\theta}}}
    \Bigg [\frac{1}{G}\sum_{g=1}^{G}\widehat{A}^{(g)}_t\log \pi_{\theta_i}(a_{i,t}^{(g)}|h_{i,t}^\mathcal{G}) \Bigg].
    \label{eq:calculate_gradient}
\end{equation}

\section{Experiments}

We evaluate MAGRPO on LLM writing and coding collaboration. Datasets, reward specifications, and additional results are provided in the Appendix.

\subsection{Writing Collaboration}

\begin{table*}[t]
\footnotesize
\centering
\begin{tabular}{lccccccc}
\toprule
\multirow{2}{*}{\textbf{Method}} & \multirow{2}{*}{\textbf{Dataset}} & \multicolumn{2}{c}{\textbf{Efficiency}} & \multicolumn{3}{c}{\textbf{Article Quality (\%)}} & \multirow{2}{*}{\textbf{Return (\%)}}\\
\cmidrule(lr){3-4} \cmidrule(lr){5-7}
 & & \textbf{Speed} & \textbf{Response Time} & \textbf{Structure} & \textbf{Consistency} & \textbf{Coherence}  \\
\midrule
\multirow{2}{*}{\text{Single Model}} & TLDR & 64.1 & 6.6 & 43.8 & 97.6 & 52.8 & 36.7 \\
 & arXiv & 65.4 & 6.5 & 51.2 & 87.2 & \textbf{71.1} & 44.9 \\
\multirow{2}{*}{\text{Parallel Generation}} & TLDR & 185.6 & \textbf{2.1} & 25.9 & 98.3 & 56.5 & 23.2 \\
 & arXiv & 190.6 & \textbf{2.1} & 71.5 & 64.2 & 61.5 & 59.6 \\
\multirow{2}{*}{\text{Sequential Generation}} & TLDR & 98.7 & 4.3 & 33.5 & 98.5 & 64.5 & 21.7 \\
 & arXiv & 85.8 & 4.3 & 92.4 & \textbf{97.8} & 64.3 & 87.7 \\
\multirow{2}{*}{\text{One-Round Discussion}} & TLDR & 100.4 & 4.3 & 35.9 & \textbf{98.8} & 60.8 & 22.3 \\
 & arXiv & 95.4 & 4.3 & 84.6 & 71.8 & 66.0 & 76.6 \\
\midrule
\multirow{2}{*}{\text{MAGRPO (Ours)}} & TLDR & \textbf{202.3} & \textbf{2.1} & \textbf{98.7} & 97.1 & \textbf{78.5} & \textbf{94.5} \\
 & arXiv & \textbf{193.8} & \textbf{2.1} & \textbf{97.9} & 96.2 & 69.7 & \textbf{93.1} \\
\bottomrule
\end{tabular}
\caption{Performance of MAGRPO against baselines on TLDR and arXiv. Speed (tokens/s) and response time (s) are measured on GeForce RTX 5090s. Results are normalized to the return scale. \textbf{Bolds} indicate the best performance on each dataset.}
\label{tab:article_generation}
\end{table*}

We explore LLM collaboration for article writing using MAGRPO across 2 classic tasks: summarization and expansion.

\begin{figure*}[t]
    \centering
    \hspace{6mm}
    \begin{subfigure}[b]{0.32\textwidth}
        \centering
        \includegraphics[width=\textwidth]{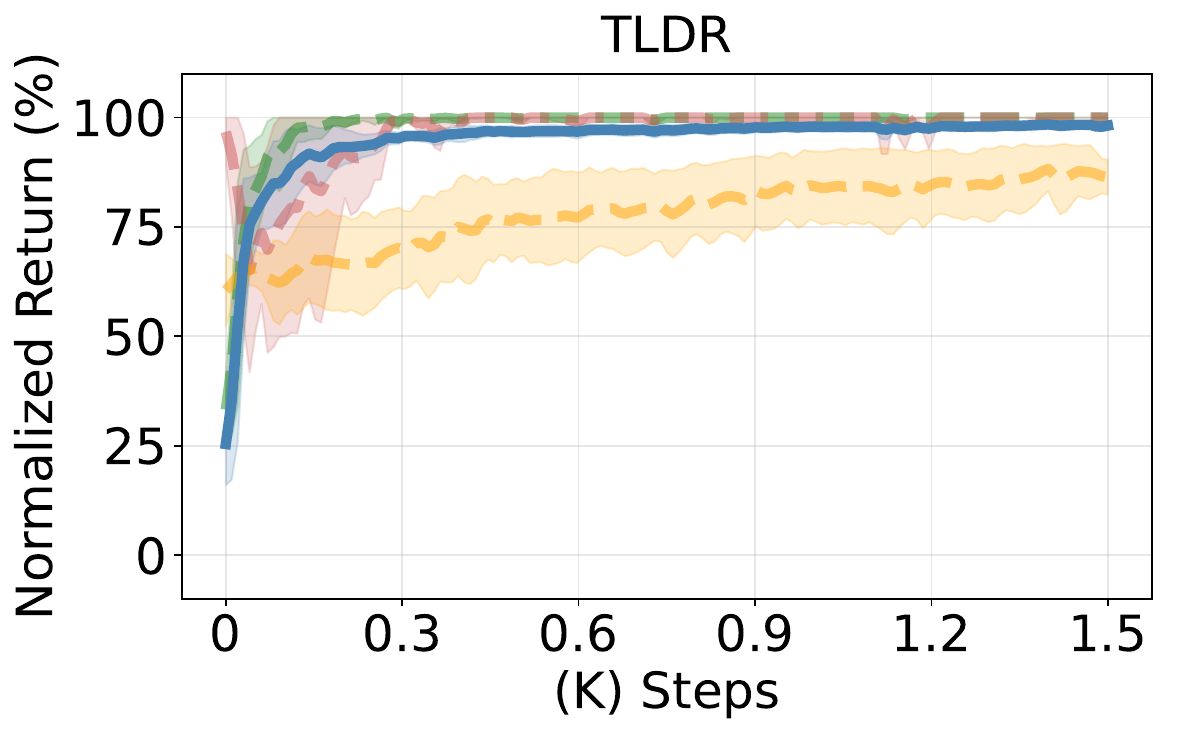}
        \caption{TLDR summarization}
        \label{fig:tldr_curves}
    \end{subfigure}
    \hspace{4mm}
    \begin{subfigure}[b]{0.48\textwidth}
        \centering
        \includegraphics[width=\textwidth]{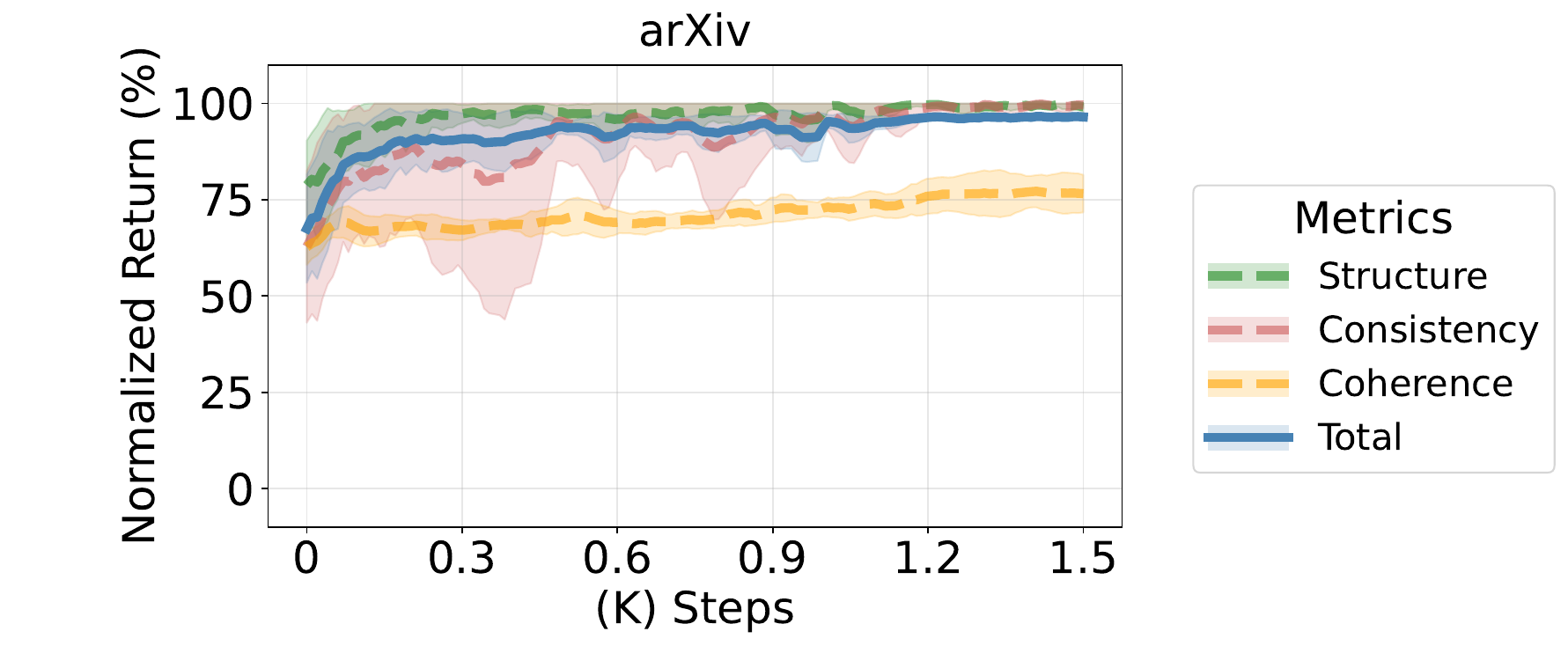}
        \caption{arXiv abstract expansion}
        \label{fig:arxiv_curves}
    \end{subfigure}
    \caption{Normalized returns on writing collaboration: (a) structural wellness (dashed green); (b) style consistency (dashed red); (c) coherence (dashed orange); (d) total rewards (solid blue). All returns are normalized to the return scale.}
    \label{fig:ag_training_curves}
\end{figure*}

\subsubsection{TLDR Summarization}

When reading a long article, readers often seek to quickly grasp its core ideas. If the topic is of interest, they may wish to delve deeper into specific details while still avoiding a complete reading through the full document. This calls for a summarization system to generate summaries at varying levels of detail. We frame this task using TLDR summarization as an illustrative example.

The TLDR dataset comprises unabridged Reddit posts in the \texttt{prompt} and concise summaries appended by the author in the \texttt{completion}. In our experiment, 2 \textit{Qwen3-1.7B} agents independently summarize the \texttt{prompt} without using \texttt{completion}. The first agent functions as a core-idea (TLDR) generator, producing a concise paragraph, while the second agent serves as a detailed summarizer, providing more comprehensive information. 

To quantify the summarization quality, we employ a relatively simple combination of 3 metrics.  Structure measures the lengths and the length ratio of the 2 summaries, to ensure the TLDR is concise and the detailed summary is sufficiently long. Style consistency is assessed using the normalized Jaccard similarity coefficient, calculated as the ratio of the intersection size to the union size of unique words (or n-grams) between responses. A high style consistency reward typically indicates that the summarizers adopt similar stylistic patterns while avoiding identical wording. Logical coherence is quantified by counting the occurrences of transition words. Positive reward is given for using transition words, but the reward decreases logarithmically as more are used. These metrics are simple approximations of what more complex reward models may evaluate. Other (simpler or more complex) metrics or reward models could also be used. The total reward combines these metrics through a weighted summation. More details regarding our reward model and hyperparameters are provided in the Appendix.

\subsubsection{arXiv Expansion}

Writing a long article typically requires contributions from multiple writers, each responsible for different sections. As a simple scenario, 2 agents can collaborate to generate introductions from the \texttt{abstract} of arXiv papers. The first agent outlines the research background and motivation, while the other presents the proposed methods and their experiments. The combined paragraphs should be coherent and consistent in style. Similar to the reward model in TLDR summarization, we employ the same evaluation metrics as proxies, with threshold hyperparameters specifically adjusted for this task.

\subsubsection{Baselines} We adopt a single model and 3 multi-agent methods as our baselines. To minimize the influence of prompts on our comparison, we keep the task description fixed and only add minimal instructions. In the single-agent baseline, we prompt with the article, the agent's role (summarizer or expanding writer), and specific format instructions. Naive concatenation builds on it by dividing the task into subtasks, assigning each agent a specific portion to complete in parallel without explicit communication. The sequential pipeline introduces one-way communication, allowing one agent to respond based on both the task description and the other agent’s output. The one-round discussion baseline enables bidirectional communication: agents first receive the same prompts as in naive concatenation, then the prompts are augmented with the other’s first-turn response in the second turn. All baseline methods operate without fine-tuning and depend solely on prompt-level interactions. Detailed prompts for each baseline are in the Appendix. 

\begin{figure*}[htb]
    \centering
    \hspace{6mm}
    \begin{subfigure}[b]{0.32\textwidth}
        \centering
        \includegraphics[width=\textwidth]{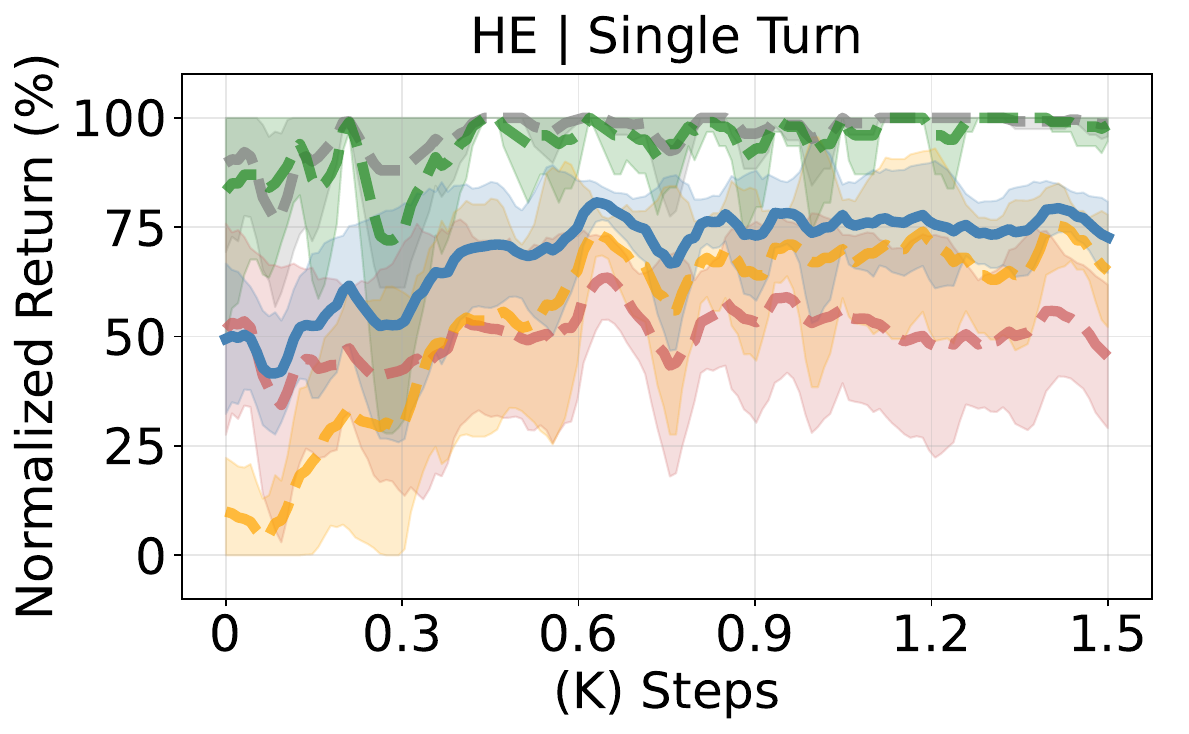}
        \caption{Single-Turn MAGRPO on HE}
        \label{fig:he_one_turn}
    \end{subfigure}
    \hspace{4mm}
    \begin{subfigure}[b]{0.48\textwidth}
        \centering
        \includegraphics[width=\textwidth]{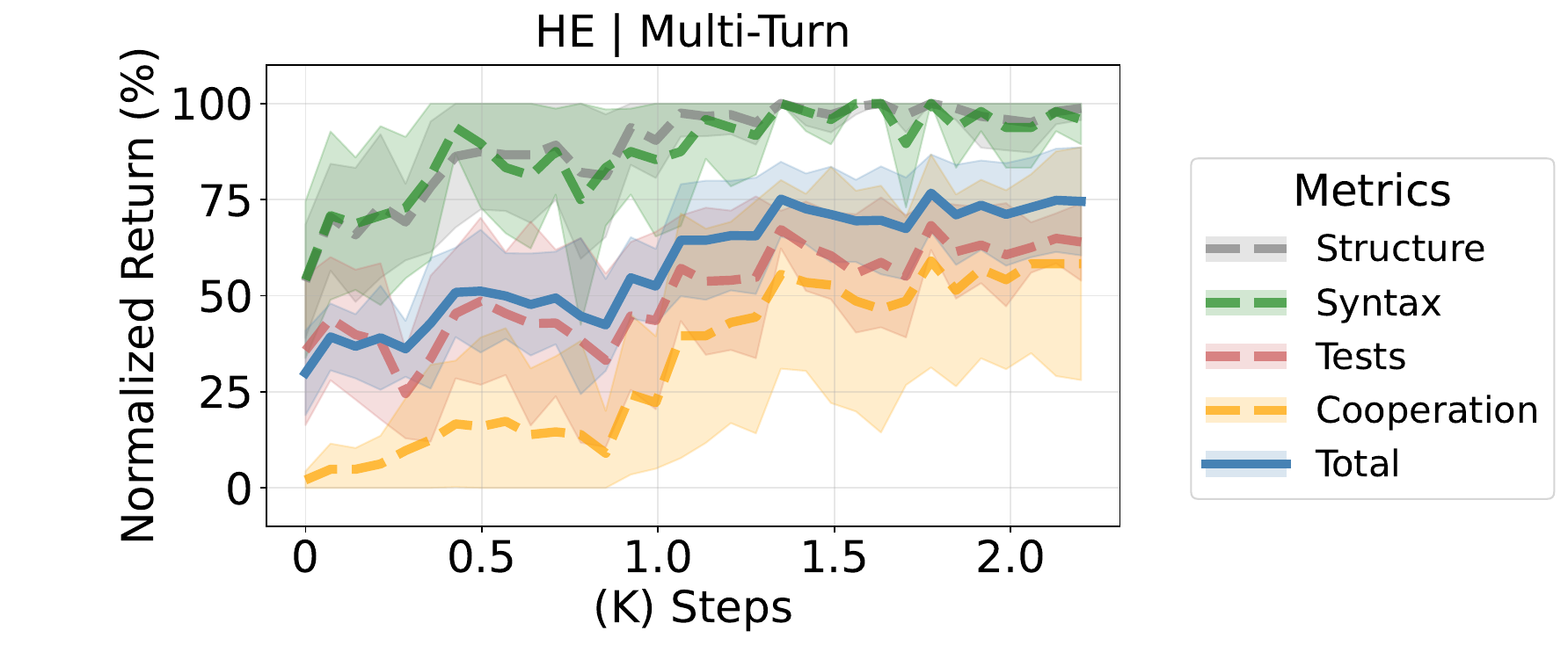}
        \caption{Multi-Turn MAGRPO on HE}
        \label{fig:he_2_turn}
    \end{subfigure}

    \hspace{6mm}
    \begin{subfigure}[b]{0.32\textwidth}
        \centering
        \includegraphics[width=\textwidth]{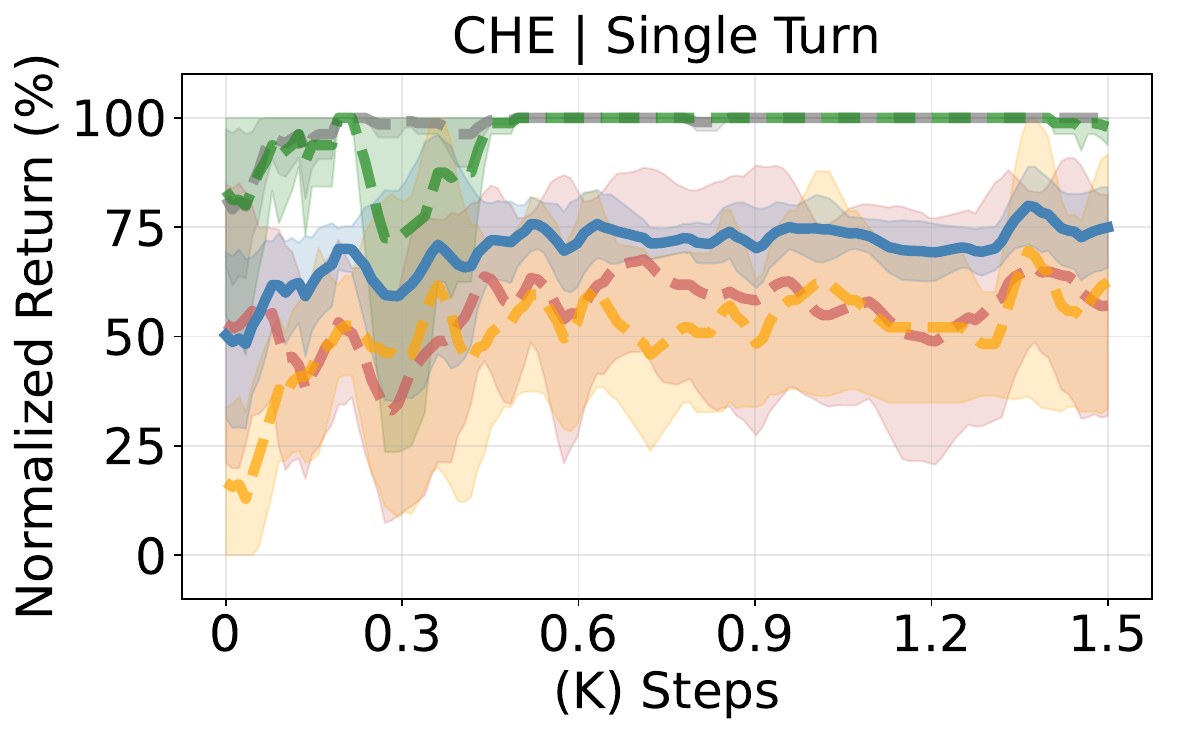}
        \caption{Single-Turn MAGRPO on CHE}
        \label{fig:che_one_turn}
    \end{subfigure}
    \hspace{4mm}
    \begin{subfigure}[b]{0.48\textwidth}
        \centering
        \includegraphics[width=\textwidth]{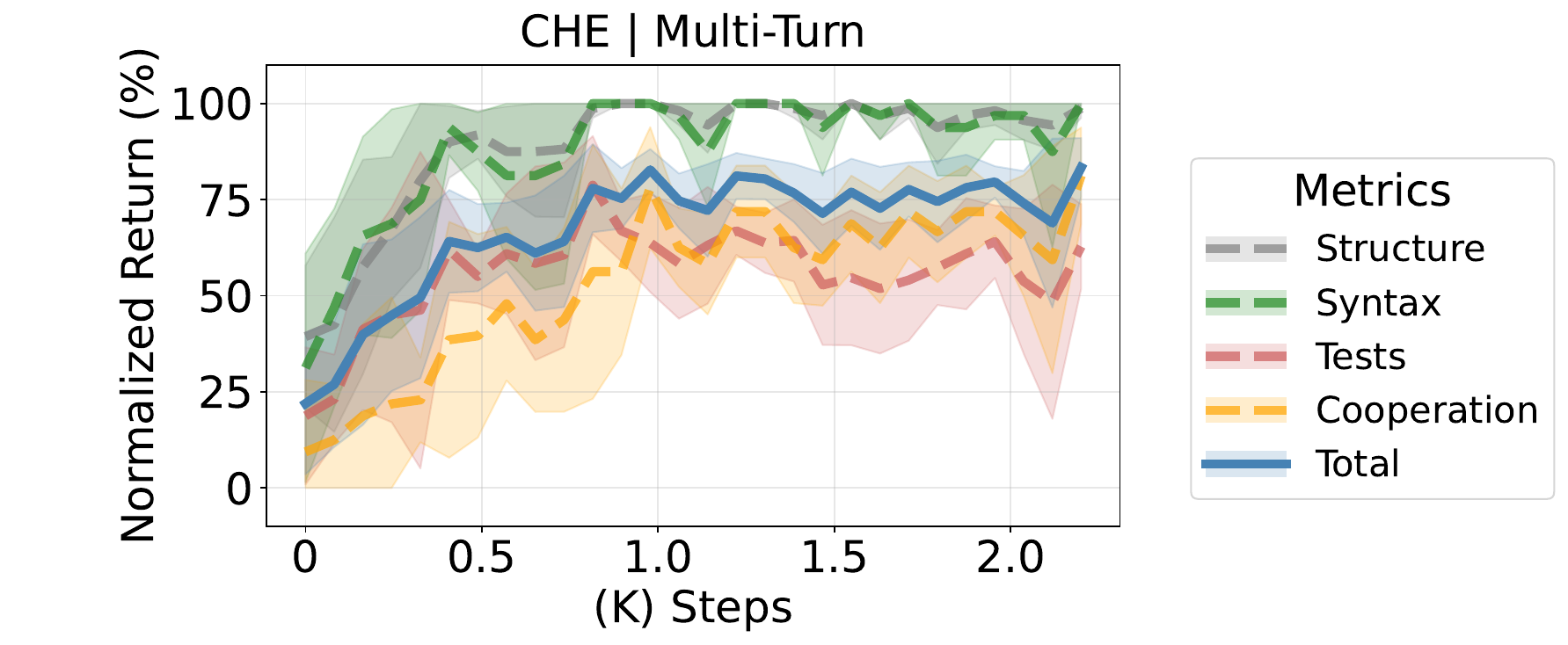}
        \caption{Multi-Turn MAGRPO on CHE}
        \label{fig:che_2_turn}
    \end{subfigure}
    \caption{Normalized returns on coding collaboration: (a) structural wellness (dashed grey); (b) syntax correctness (dashed green); (c) Test score (dashed red); (d) cooperation rewards (dashed yellow); (e) total return (solid blue).}
    \label{fig:code_training_curves}
\end{figure*}
\begin{table*}[t]
\footnotesize
\centering
\begin{tabular}{lcccccccc}
\toprule
\multirow{2}{*}{\textbf{Method}} & \multirow{2}{*}{\textbf{Dataset}}
  & \multicolumn{2}{c}{\textbf{Efficiency}}
  & \multicolumn{4}{c}{\textbf{Code Quality (\%)}} & \multirow{2}{*}{\textbf{Return  (\%)}} \\
\cmidrule(lr){3-4} \cmidrule(lr){5-8}
 & 
  & \textbf{Speed} & \textbf{Response Time}
  & \textbf{Structure} & \textbf{Syntax} & \textbf{Tests}
  & \textbf{Cooperation} \\
\midrule
\multirow{2}{*}{\text{Single Model}}
  & HE  & 73.1 & 1.6 & \textbf{100.0} & \textbf{100.0} & 64.8 & -- & -- \\
  & CHE & 65.5 & 1.4 & \textbf{100.0} & \textbf{100.0} & 63.4 & -- & -- \\
\multirow{2}{*}{\text{Fine-Tuned Single Model}}
  & HE  & 72.2 & 1.6 & \textbf{100.0} & 99.2 & 65.6 & -- & -- \\
  & CHE & 60.6 & 1.4 & \textbf{100.0} & \textbf{100.0} & 66.7 & -- & -- \\
\multirow{2}{*}{\text{Naive Concatenation}}
  & HE  & \textbf{194.9} & \textbf{1.1} & 96.1 & 90.6 & 42.5 & 22.7 & 53.9 \\
  & CHE & 189.4 & \textbf{1.1} & 97.5 & 95.0 & 40.1 & 24.0 & 54.3 \\
\multirow{2}{*}{\text{Sequential Pipeline}}
  & HE  & 99.6 & 2.2 & 98.4 & 96.5 & 56.4 & 35.1 & 63.1 \\
  & CHE & 97.4 & 2.0 & 97.5 & 96.3 & 55.2 & 35.2 & 62.5 \\
\multirow{2}{*}{\text{One-Round Discussion}}
  & HE  & 82.5 & 2.8 & 98.1 & 94.8 & 41.2 & 30.2 & 57.5 \\
  & CHE & 78.3 & 2.8 & 97.5 & 96.3 & 41.9 & 34.8 & 59.5 \\
\midrule
\multirow{2}{*}{\text{Single-Turn MAGRPO (Ours)}}
  & HE  & 190.0 & 1.5 & \textbf{100.0} & 97.8 & 61.6 & 83.4 & 83.7 \\
  & CHE & \textbf{192.4} & 1.5 & 98.8 & 97.5 & 71.2 & 83.7 & 86.0 \\
\multirow{2}{*}{\text{Multi-Turn MAGRPO (Ours)}}
  & HE  & 96.5 & 2.7
         & 99.7 & 96.9 & \textbf{68.4}
         & \textbf{84.9} & \textbf{86.7} \\
  & CHE & 97.1 & 2.5
         & 98.6 & 98.2 & \textbf{75.0}
         & \textbf{86.3} & \textbf{88.5} \\
\bottomrule
\end{tabular}
\caption{Performance comparison of MAGRPO against baselines on HE and CHE. Speed (tokens/s) and response time (s) are recorded on GeForce RTX 5090s. Results are normalized to the return scale and averaged over 10 runs; rewards are level-based. \textbf{Bolds} indicate the best performance of each metric on each dataset.}\label{tab:coder_generation}
\end{table*}

\subsubsection{Results}

In this experiment, we apply MAGRPO to optimize the dual \textit{Qwen3-1.7B} system in one turn. Figure~\ref{fig:tldr_curves} and Figure~\ref{fig:arxiv_curves} show the evaluation results on TLDR and arXiv over 10 runs. The upward trend on all metric curves indicates that 2 agents gradually cooperate to generate coherent and consistent content with a well-organized structure. In the TLDR summarization, while the structure and logical coherence monotonically increase throughout training, the style consistency curves exhibit a decrease in the first 100 steps. This occurs as agents temporally diverge in styles to optimize other cooperative objectives, but their styles are gradually realigned and stabilized with sufficient training. 

As shown in Table~\ref{tab:article_generation}, MAGRPO is 3 times faster compared to the single \textit{Qwen3-4B} model, which has a comparable number of parameters to our dual \textit{Qwen3-1.7B} system. Despite receiving detailed instructions, \textit{Qwen3-4B} fails to produce well-structured responses. A similar issue appears in TLDR summarization but not in arXiv expansion under multi-agent settings. This is because the outputs of homogeneous agents are naturally similar in length, which fortuitously aligns with the preference of the reward model.

Among the multi-agent baselines, parallel generation is the only one that achieves a comparable speed to ours, but it fails to generate well-structured and coherent texts due to the lack of cooperation. Sequential generation and discussion-based approaches occasionally enhance coordination through specific prompts. However, they still underperform ours in efficiency and coherence, resulting in lower total return. The limited effectiveness of prompt-instructed coordination constrains their scalability to more complex scenarios involving large numbers of agents or extended multi-turn interactions \cite{debatefail}.

\subsection{Coding Collaboration}

In large-scale software development, numerous developers collaborate to implement complex systems. Employing LLMs as developers is a promising direction, but coordinating them is challenging due to diverse cooperation schemes and complex failure modes. We simplify this task by using 2 \textit{Qwen2.5-Coder-3B} agents to generate Python functions collaboratively. A helper agent produces auxiliary functions to support a main function generator, without any direct communication. The outputs from both agents, along with required libraries, are aggregated into complete code snippets.

\subsubsection{HumanEval}
We evaluate MAGRPO on the HumanEval (HE) dataset, which contains 164 handwritten programming problems, each containing a natural language description (\texttt{prompt}), a function signature (\texttt{entry\_point}), and a set of unit tests (\texttt{test}). To guide learning, we design a level-based reward model that prioritizes fundamental aspects of code generation. Structural integrity verifies the presence and correctness of both main and auxiliary function definitions; syntactic correctness ensures compliance with Python syntax; test pass rate assesses functional correctness based on the proportion of successfully passed unit tests; and a cooperation quality bonus is granted when the main function properly invokes and utilizes the auxiliary function. Rewards are accumulated only when all requirements at each preceding level are satisfied.

\subsubsection{CoopHumanEval} 
Some entries in HumanEval (HE) are not designed for coding collaboration; certain atomic operations (e.g., \texttt{strlen(string)}) can hardly be decomposed in a way that has meaningful cooperation. These noisy instances bias it towards invalid cooperation, such as merely wrapping the auxiliary function, thereby making training unstable. Thus, we construct a cooperative code generation dataset, CoopHumanEval (CHE), which comprises both original HE problems with cooperative potential (e.g., \texttt{prime\_fib(n)}) and additional problems (e.g., \texttt{unique\_digits(x)}). CHE problems are decomposable, where agents can explore effective cooperation schemes.

\subsubsection{Baselines}
We adopt the fixed and fine-tuned single model, and 3 multi-agent methods on fixed base models, as our baselines. In the single-agent setting, the \textit{Qwen2.5-Coder-7B} model generates a Python function based on the problem description in \texttt{prompt}. We also fine-tune this model on the training set to adapt it to this task. In the multi-agent setting, 2 \textit{Qwen2.5-Coder-3B} models serve as agents: one generates a helper function, and the other produces the main. To minimize the influence of prompts on our comparison, we keep the problem description fixed and only add minimal coordination instructions. In the naive concatenation, agents are informed of their roles and generate outputs in parallel without communication. The sequential pipeline allows the main agent to respond based on the other's output. In the one-round discussion, agents first receive the same prompts as naive concatenation, then the prompts are augmented with the other’s last-turn response in the subsequent turns.

\subsubsection{Results} We optimize the interaction between 2 agents using single-turn and multi-turn MAGRPO. In the multi-turn setting, problem descriptions and the agents’ initial responses are provided to static checking (\textit{AST}) and dynamic execution models, which report errors for each agent. Ablations of external tools used, including self-evolving and expert guidance, are provided in the Appendix.

Figure~\ref{fig:he_one_turn} and Figure~\ref{fig:he_2_turn} show the normalized return of MAGRPO on HE over 10 runs. Single-turn MAGRPO training improves the syntactical correctness and develops valid cooperation, while the test pass rate does not show much progress. As for the multi-turning training, agents are initially overwhelmed by the external model's feedback, resulting in even lower initial returns. They gradually adopt the error signals and improve their returns. However, the test pass rate shows no significant improvement over the single-agent model, due to noisy entries in the dataset and hence unreliable feedback. This reflects the complexity and delicacy of decentralized coder coordination, where the main agent must accurately infer the functionality of auxiliary modules and trust their correctness without communication.

The performance of single-turn and multi-turn MAGRPO on the CHE dataset is shown in Figures~\ref{fig:che_one_turn} and~\ref{fig:che_2_turn}. Results show that MAGRPO achieves higher overall returns and lower variance when trained on CHE over HE. In the multi-turn setting, although agents initially struggle to interpret the feedback, like training on HE, the returns gradually improve and eventually surpass those in the single-turn training. This demonstrates that, when trained on a dataset with well-defined cooperative structures, agents can learn to utilize error messages to improve their response quality.

Table~\ref{tab:coder_generation} presents a performance comparison between MAGRPO and baselines on both HE and CHE. By GRPO fine-tuning, the performance of \textit{Qwen2.5-Coder-7B} model only improves slightly as the logic of test problems differs substantially. Compared to a single model, the naive concatenation method has lower test pass rates, as the main agent may generate code based on incorrect assumptions about auxiliary functionality. In the sequential pipeline method, the main agent can provide a backup for the auxiliary function when it identifies potential vulnerabilities, thereby improving the test pass rate. However, this comes at the cost of slower inference speed. Although the one-round discussion method involves more communication between agents, its effectiveness remains limited because the agents’ mutual adaptations to each other’s last responses can become misaligned. MAGRPO outperforms all baselines on both CHE and HE by facilitating effective cooperation and leveraging feedback from the external model. Additional results, including pass@k, are presented in the Appendix.

\subsubsection{Cooperation Schemes} MAGRPO identifies diverse cooperation schemes. In some cases, the auxiliary function handles the core logic, while the main agent adds backup logic or decorations to improve the overall solution. Alternatively, the main agent may act as a coordinator, decomposing the problem and assigning subtasks to the auxiliary agent. The auxiliary function may serve as a strategy filter, guiding the main agent to generate code for specific cases. While coordinator and strategy-filter schemes can improve inference efficiency, they are more prone to syntax and logical errors. With limited cooperation-oriented training data, the main agent typically resorts to more conservative roles, i.e., fallback or decoration. These cooperation schemes emerge during training under a relatively simple joint reward. More refined design patterns will likely be found when training agents to develop large-scale coding projects.

\section{Conclusion}

In this paper, we model LLM collaboration as a cooperative MARL problem and formalize it as a Dec-POMDP. We propose the MAGRPO algorithm to optimize agent cooperation through shared rewards. Our experiments in coding and writing collaboration show that MAGRPO enables agents to efficiently generate high-quality responses via effective collaboration. Our work encourages future exploration of MARL-based methods for scalable and robust LLM collaboration.

\section{Acknowledgments}

This work was partially funded by NSF grants \#2044993 and \#2409351. It used Delta and DeltaAI computing resources at the National Center for Supercomputing Applications through allocation CIS250443 and CIS250554 from the Advanced Cyberinfrastructure Coordination Ecosystem: Services \& Support program, which is supported by NSF grants \#2138259, \#2138286, \#2138307, \#2137603, and \#2138296. 

We thank Tianle Chen for improving our MARL training framework for LLM collaboration, CoMLRL, and members of the Lab for Learning and Planning in Robotics (LLPR) for the valuable discussion. We thank Gregory Bauer and Brett Bode for helping us resolve job failure issues. 

\bibliography{aaai2026}

\appendix

\section{Formalizations of Multi-Agent Interaction}

Many studies adopt Partially Observable Stochastic Games (POSG) to model the LLM interaction in MAS \cite{spiral, amongus, marti2025}. In this section, we show that Dec-POMDP offers special merits compared to POSG in the solution concept in the cooperative settings, thus more suited to model LLM collaboration.

\subsection{Dec-POMDP} 

A Dec-POMDP is defined by $\langle \mathcal{I}, \mathcal{S}, \{\mathcal{O}_i\}, \{\mathcal{A}_i\}, R, T, H \rangle$. At each step $t$, since an agent cannot directly observe the state $s_t$, it usually maintains local observation-action history $h_{i, t} = (o_{i, 0}, a_{i, 0}, \ldots, o_{i, t})$ to infer a belief over the underlying state. Decisions are made according to a local policy $\pi_i: \mathcal{H}_{i,t} \rightarrow \Delta(\mathcal{A}_i)$, which maps histories to probability distributions over actions. The set of all local policies forms the joint policy $\boldsymbol{\pi} = \{\pi_1, \ldots, \pi_n\}$. In cooperative settings, the objective is to maximize shared cumulative rewards. As proved in \cite{fransoptimalq}, there is always an optimal joint policy in a Dec-POMDP,
\begin{equation}
\boldsymbol{\pi}^*=\argmax_{\boldsymbol{\pi}\in \boldsymbol{\Pi}} \mathbb{E}_{\boldsymbol{\pi}}\left[\sum_{t=0}^{H-1} R(s_t, a_t) \right].
\end{equation}

\subsection{POSG} 

A Partially Observable Stochastic Game (POSG), so-called Partially Observable Markov Game (POMG), does not assume cooperative behavior among agents. It can be either a cooperative, competitive, or mixed game. A POSG is defined as $\langle \mathcal{I}, \mathcal{S}, \{\mathcal{A}_i\}, T, \{\mathcal{O}_i\}, O, \{R_i\}, H \rangle$, where each agent has its own reward function $R_i: \mathcal{S} \times \mathcal{A} \rightarrow \mathbb{R}$. In POSG, each agent seeks to maximize its individual return under the fixed policies of all others $\boldsymbol{\pi}_{-i}$. The optimal policy $\pi_i^\circledast$ for each agent $i\in\mathcal{I}$ is,
\begin{equation}
\pi_i^\circledast=\argmax_{\pi_i\in \Pi_i} \mathbb{E}_{\pi_i,\boldsymbol{\pi}_{-i}}\left[\sum_{t=0}^{H-1} R_i(s_t, a_t)\right],
\end{equation}
The solutions for POSG are Nash Equilibria (NE), where no agents can unilaterally improve their returns by deviating from their policies. Formally, for all $i \in \mathcal{I}$ and any alternative policy $\pi_{i} \in \Pi_i$, NE satisfy
\begin{equation}
\mathbb{E}\left[\sum_{t=0}^{H-1} R_i(s_t, a_t) \mid \pi^\circledast_{i}, \boldsymbol{\pi}^\circledast_{-i}\right] \geq \mathbb{E}\left[\sum_{t=0}^{H-1} R_i(s_t, a_t) \mid \pi_i, \boldsymbol{\pi}^\circledast_{-i}\right].
\end{equation}

Like Dec-POMDP, the decision-making in POSG is still concurrent (as stochastic games), where all agents act synchronously at each time step. In contrast, turn-based interactions, where agents take turns to act (e.g., chess, Kuhn Poker, tic-tac-toe), are typically modeled as extensive-form games.

\subsection{Non-Optimality of POSG Solutions}

We illustrate that the solutions of POSG, i.e., NE, may not necessarily lead to joint optimality in cooperative settings.

Consider a one-step matrix game involving 2 agents, where each agent selects an action from the action space $\mathcal{A}=\{\mathcal{A}^{(1)}, \mathcal{A}^{(2)}\}$. The joint action profile determines the utility as presented in Table~\ref{tab:joint_payoff}.

\begin{table}[H]
\centering
\begin{subtable}{0.22\textwidth}
\centering
\begin{tabular}{c|cc}
$a_1 \backslash a_2$ & $\mathcal{A}^{(1)}$ & $\mathcal{A}^{(2)}$ \\
\hline
$\mathcal{A}^{(1)}$ & 10 & 7 \\
$\mathcal{A}^{(2)}$ & 7 & 0
\end{tabular}
\end{subtable}
\caption{Joint utility matrix of 2 agents.}
\label{tab:joint_payoff}
\end{table}

This matrix game can be potentially decomposed into 2 POSG in Table~\ref{tab:posg} through reward shaping.

\begin{table}[H]
\centering
\begin{subtable}{0.22\textwidth}
\centering
\begin{tabular}{c|cc}
$a_1 \backslash a_2$ & $\mathcal{A}^{(1)}$ & $\mathcal{A}^{(2)}$ \\
\hline
$\mathcal{A}^{(1)}$ & (5, 5) & (3, 4) \\
$\mathcal{A}^{(2)}$ & (4, 3) & (0, 0)
\end{tabular}
\caption{POSG 1}
\label{tab:posg1}
\end{subtable}
\hfill
\begin{subtable}{0.22\textwidth}
\centering
\begin{tabular}{c|cc}
$a_1 \backslash a_2$ & $\mathcal{A}^{(1)}$ & $\mathcal{A}^{(2)}$ \\
\hline
$\mathcal{A}^{(1)}$ & (5, 5) & (1, 6) \\
$\mathcal{A}^{(2)}$ & (6, 1) & (0, 0)
\end{tabular}
\caption{POSG 2}
\label{tab:posg2}
\end{subtable}
\caption{Return tables of 2 POSG.}
\label{tab:posg}
\end{table}

\begin{figure}[H]
    \centering
    \begin{subfigure}{0.22\textwidth}
        \centering
        \includegraphics[width=\linewidth]{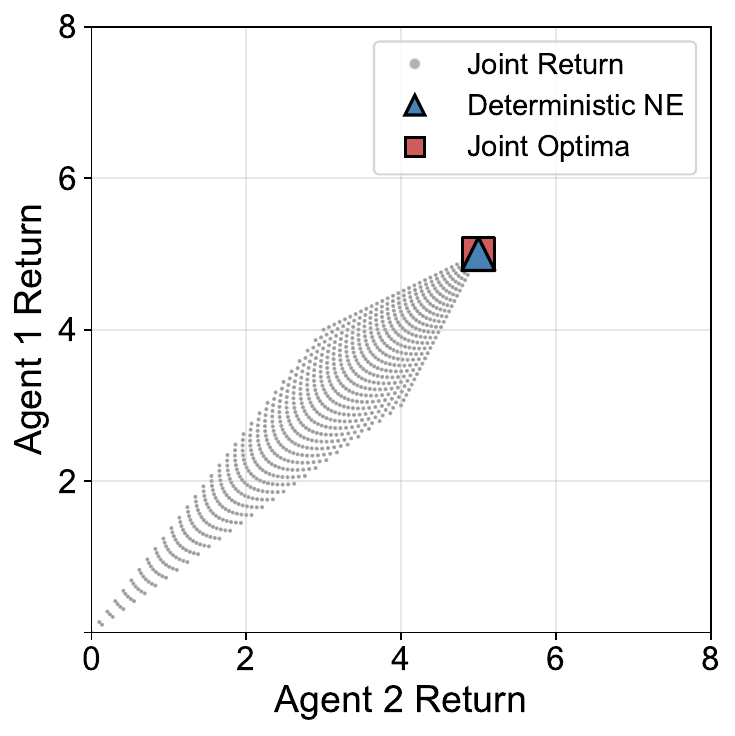}
        \caption{POSG 1}
        \label{fig:posg1}
    \end{subfigure}
    \hspace{3mm}
    \begin{subfigure}{0.22\textwidth}
        \centering
        \includegraphics[width=\linewidth]{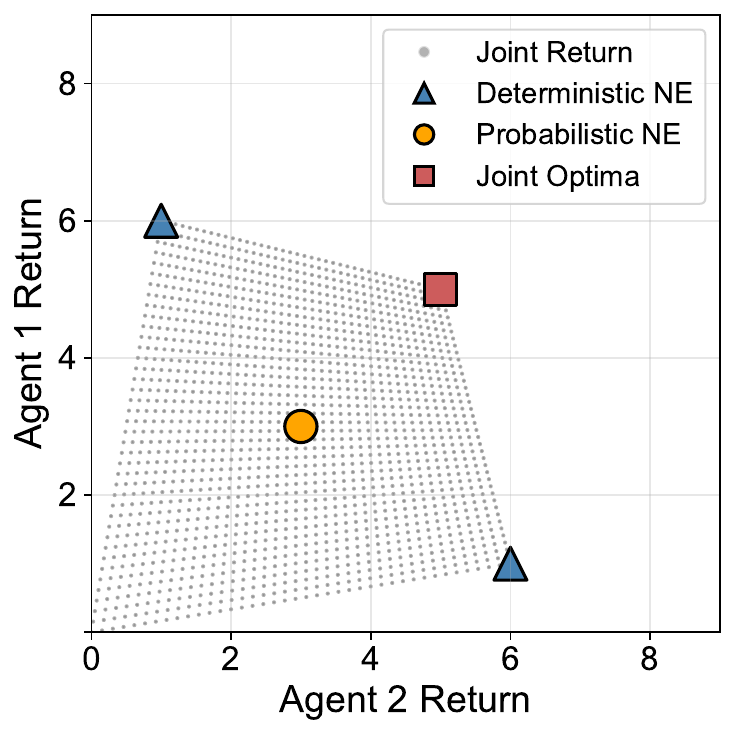}
        \caption{POSG 2}
        \label{fig:posg2}
    \end{subfigure}
    \caption{Utility spaces of 2 POSG.}
    \label{fig:posg}
\end{figure}

In the POSG presented in Table~\ref{tab:posg1}, $(\mathcal{A}^{(1)}, \mathcal{A}^{(1)})$ is a Nash equilibrium (blue triangle in Figure~\ref{fig:posg1}). When $a_1=\mathcal{A}^{(1)}$, $U_2(\mathcal{A}^{(1)}, \mathcal{A}^{(1)}) > U_2(\mathcal{A}^{(1)}, \mathcal{A}^{(2)})$; when $a_1=\mathcal{A}^{(2)}$, $U_2(\mathcal{A}^{(2)}, \mathcal{A}^{(1)}) > U_2(\mathcal{A}^{(2)}, \mathcal{A}^{(2)})$. Therefore, the best response for agent 2 is $a_2^\circledast=\mathcal{A}^{(1)}$. Similarly, since $U_1(\mathcal{A}^{(1)}, \mathcal{A}^{(1)}) > U_1(\mathcal{A}^{(2)}, \mathcal{A}^{(1)})$, we obtain $a_1^\circledast=\mathcal{A}^{(1)}$. This NE also achieves joint optimality with the maximum utility $5+5=10$ (red square in Figure~\ref{fig:posg1}).

\begin{table*}[t]
\centering
\footnotesize
\begin{tabular}{lc|ccc|ccc|ccc}
\toprule
\multirow{2}{*}{\textbf{Method}}& \multirow{2}{*}{\textbf{Dataset}}& \multicolumn{3}{c|}{\textbf{Pass@k (\%)}} & \multicolumn{3}{c|}{\textbf{Acc@k} (\%)} & \multicolumn{3}{c}{\textbf{Coop@k} (\%)} \\
 & 
  & \textbf{@3} & \textbf{@5} & \textbf{@10}
  & \textbf{@3} & \textbf{@5} & \textbf{@10}
  & \textbf{@3} & \textbf{@5} & \textbf{@10} \\
\midrule
\multirow{2}{*}{\textbf{Single Model}}
  & HE  & 67.7 & 71.0 & 83.9 & 85.4 & 87.9 & 95.1 & -- & -- & -- \\
  & CHE & 68.8 & 75.0 & 81.3 & 75.0 & 81.3 & 88.8 & -- & -- & -- \\
\multirow{2}{*}{\textbf{Fine-Tuned Single Model}}
  & HE  & 69.2 & 72.9 & 84.6 & 83.0 & 89.2 & \textbf{95.6} & -- & -- & -- \\
  & CHE & 69.4 & 74.5 & 83.7 & 74.3 & 82.2 & 90.3 & -- & -- & -- \\
\multirow{2}{*}{\textbf{Naive Concatenation}}
  & HE  & 45.2 & 51.6 & 64.5 & 70.4 & 75.5 & 80.9 & 49.5 & 67.7 & 76.3 \\
  & CHE & 43.8 & 56.3 & 68.8 & 57.0 & 63.8 & 73.8 & 47.9 & 69.3 & 81.3 \\
\multirow{2}{*}{\textbf{Sequential Pipeline}}
  & HE  & 56.8 & 61.3 & 71.0 & 78.8 & 84.5 & 91.9 & 62.4 & 73.3 & 92.7 \\
  & CHE & \textbf{75.0} & \textbf{81.5} & \textbf{87.5} & \textbf{88.2} & 89.5 & 91.3 & 75.0 & 75.0 & 81.3 \\
\multirow{2}{*}{\textbf{One-Round Discussion}}
  & HE  & 51.6 & 61.3 & 71.0 & 71.8 & 81.3 & 87.5 & 58.1 & 70.0 & 78.7 \\
  & CHE & 50.0 & 68.8 & 68.8 & 75.4 & 82.5 & 82.0 & 66.7 & 68.7 & 75.0 \\
\midrule
\multirow{2}{*}{\textbf{Single-Turn MAGRPO (Ours)}}
  & HE  & 64.8 & 68.1 & 71.0 & 75.3 & 76.3 & 86.4 & 83.8 & 90.3 & 90.3 \\
  & CHE & \textbf{75.0} & 75.0 & 81.2 & 80.0 & 82.5 & 87.5 & 87.5 & 93.8 & 93.8 \\
\multirow{2}{*}{\textbf{Multi-Turn MAGRPO (Ours)}}
  & HE  & \textbf{71.0} & \textbf{80.6} & \textbf{90.3} & \textbf{85.7} & \textbf{92.6} & 94.7 & \textbf{93.5} & \textbf{96.8} & \textbf{96.8} \\
  & CHE & \textbf{75.0} & \textbf{81.5} & \textbf{87.5} & 86.4 & \textbf{92.5} & \textbf{95.4} & \textbf{93.8} & \textbf{96.8} & \textbf{100.0} \\
\bottomrule
\end{tabular}
\caption{Performance comparison between MAGRPO and baseline methods with pass@k, acc@k, coop@k, on HE and CHE. The \textbf{bold} texts indicate the best performance of each metric on each dataset.}
\label{tab:k_metrics}
\end{table*}

However, certain reward decompositions may yield non-optimal solutions for cooperative games in Table~\ref{tab:posg}, even when POSG solutions reach NE. For the POSG shown in Table~\ref{tab:posg2}, the deterministic NE are ${(\mathcal{A}^{(1)}, \mathcal{A}^{(2)}), (\mathcal{A}^{(2)}, \mathcal{A}^{(1)})}$ (blue triangles in Figure~\ref{fig:posg2}). When $a_1=\mathcal{A}^{(1)}$, agent 2 prefers $\mathcal{A}^{(2)}$ as $U_2(\mathcal{A}^{(1)}, \mathcal{A}^{(2)}) > U_2(\mathcal{A}^{(1)}, \mathcal{A}^{(1)})$; when $a_1=\mathcal{A}^{(2)}$, agent 2 prefers $\mathcal{A}^{(1)}$ since $U_2(\mathcal{A}^{(2)}, \mathcal{A}^{(1)}) > U_2(\mathcal{A}^{(2)}, \mathcal{A}^{(2)})$. Agent 1 faces the same issue. Thus, neither agent can unilaterally improve their utilities by deviating. However, the collective utilities obtained from both policies yield $6+1=7<10$, which are suboptimal compared to the joint optimum (red square in Figure~\ref{fig:posg2}).

In Table~\ref{tab:posg2}, even the probabilistic NE under stochastic policies is still non-optimal. Suppose agent 1 selects $\mathcal{A}^{(1)}$ with probability $p$, and agent 2 selects $\mathcal{A}^{(1)}$ with probability $q$, $R_1(\mathcal{A}^{(1)}, \cdot)=5q+(1-q)=4q+1$, $R_1(\mathcal{A}^{(2)}, \cdot)=6q$, $R_1(\mathcal{A}^{(1)},\cdot)=R_1\mathcal{A}^{(2)}, \cdot)$ yields $q=0.5$; similarly, $R_2(\cdot, \mathcal{A}^{(1)})=5p+(1-p)=4p+1$, $R_2(\cdot, \mathcal{A}^{(2)})=6p$, $R_2(\mathcal{A}^{(1)},\cdot)=R_2\mathcal{A}^{(2)}, \cdot)$ yields $p=0.5$. This probabilistic NE, $\pi_1^\circledast(\mathcal{A}^{(1)})=\pi_1^\circledast(\mathcal{A}^{(2)})$, $\pi_2^\circledast(\mathcal{A}^{(1)})=\pi_2^\circledast(\mathcal{A}^{(2)})$ leads to overall utilities $3+3=6 < 10$ (orange circle in Figure~\ref{fig:posg2}).

Although appropriate reward shaping techniques can transform a cooperative game into a POSG like Table~\ref{tab:posg1} to make the NE also jointly optimal, this becomes more challenging when more agents are involved and episodes become longer. We employ Dec-POMDP to avoid the intricate reward engineering and seek the joint optimality.

\section{Additional Results}

We report additional results in this section to validate the effectiveness of our approach.

\subsection{@k Ablation}

In LLMs, single-run inference often leads to high variance. To provide a more reliable evaluation of the capacities \cite{yue2025doesreinforcementlearningreally}, we evaluate the pass, test accuracy, and cooperation at $k$ runs (pass@k, acc@k, and coop@k) on MAGRPO and baselines. These @k metrics measure the best outcome among $k$ generated solutions for each problem and are averaged over all problems in $\mathcal{D}$.


Pass@k is calculated as the probability that at least one out of $k$ generated solutions passes all test cases~\cite{evaluate}. Specifically, a set $\mathcal{P}$ of $k$ solutions is randomly sampled from a pool of $\mathcal{M}$ generated solutions. To make it consistent with other @k metrics, we express pass@k as,

\begin{equation}
\text{Pass@k} = \frac{1}{|\mathcal{D}|} \sum_{j=1}^{|\mathcal{D}|}
\mathbb{E}_{\mathcal{P} \sim \text{Sample}(\mathcal{M}, k)} \left[\max_{p\in \mathcal{P}}\mathds{1} (n_j^\text{pass}=n_j) \right],
\label{eq:pass_max_at_k}
\end{equation}
where $\mathds{1}(n_j^{\text{pass}} = n_j)$ is the indicator function that equals 1 if all test cases are passed in problem $j$ (i.e., the number of passed tests $n_j^{\text{pass}}$ equals the total number of tests $n_j$), and 0 otherwise.

However, pass@k does not offer a fine-grained assessment of solution quality, as functional correctness is represented as a binary variable. As a result, failing a single test case is treated the same as failing all in pass@k. To provide a more unbiased evaluation, we use accuracy@k, defined as the highest test accuracy among the $k$ generated solutions, averaged across all problems.
\begin{equation}
\text{Acc@k} = \frac{1}{|\mathcal{D}|} \sum_{j=1}^{|\mathcal{D}|}
\mathbb{E}_{\mathcal{P} \sim \text{Sample}(\mathcal{M}, k)} \left[\max_{p\in \mathcal{P}} \frac{n_j^\text{pass}}{n_j} \right],
\label{eq:acc_max_at_k}
\end{equation}
where $n_j^{\text{pass}}$ and $n_j$ are the number of passed and total unit tests of problem $j$, and $n_j^\text{pass}/n_j$ is the test accuracy.

Similar to acc@k, we also propose coop@k, which measures the average of the highest normalized cooperation return achieved among $k$ runs across all problems. Formally, the Coop@k is defined as,
\begin{equation}
\text{Coop@k} = \frac{1}{|\mathcal{D}|} \sum_{j=1}^{|\mathcal{D}|}
\mathbb{E}_{\mathcal{P} \sim \text{Sample}(\mathcal{M}, k)} \left[ \max_{p\in\mathcal{P}} R^\text{coop}_{p} \right],
\label{eq:cooperation_max_at_k}
\end{equation}
where $R_{p}^\text{coop}=\sum_{t=0}^{H-1} r^\text{coop}_{p,t}$ is the cooperation return over horizon $H$, and $r^\text{coop}_{p,t}$ denotes the cooperation reward obtained by solution $p$ at turn $t$.

We generate $15$ samples in $\mathcal{M}$ and evaluate all methods with $k=10$. Table~\ref{tab:k_metrics} presents the results for pass@k, acc@k, and coop@k at $k = 3, 5, 10$, comparing MAGRPO with baseline methods. As expected, the trends in @k metrics are consistent with the @1 results. By fine-tuning with GRPO, the pass@k, acc@k, and coop@k are slightly improved. The naive concatenation method remains worse than the single-agent baseline in terms of pass@k, as the main agent may generate code based on incorrect assumptions about the auxiliary function. The sequential pipeline mitigates this issue by allowing the main agent to reference the auxiliary output during generation, yielding substantial improvements across all @k metrics, particularly on CHE. Although the one-round discussion method shares the same prompts as naive concatenation in the first turn, the additional discussion round yields limited improvement across the @k metrics due to misalignment in cross-adaptation.

Multi-turn MAGRPO outperforms all baselines across most @k metrics by leveraging feedback from external sources. These more comprehensive experiments further validate the general effectiveness of our approach and highlight promising directions for future work, such as training LLMs through interactions with static analyzers, sandbox-based evaluations, or expert models. Notably, the acc@k metric offers a more fine-grained perspective on performance trends as $k$ varies. In some cases, slightly increasing $k$ may not yield additional solutions that pass all test cases (as measured by pass@k), yet the improvement is still captured through higher accuracy.

\subsection{Cooperation Schemes}

By training the auxiliary and main coders to cooperate under minimal constraints (with only the problem description and their respective roles provided), diverse cooperation schemes naturally emerge. We present 4 representative schemes observed in our models.

\subsubsection{Fallback}
A commonly observed cooperation scheme is the main agent providing a fallback for the auxiliary function. Although prompted to use the auxiliary output, the main agent may still doubt its completeness and reimplement the solution as a backup in case of potential errors.

For example, the problem \texttt{compare\_one(a,b)} asks to return the larger of 2 values in its original type (string, float, or int), or \texttt{None} if equal. Listing~\ref{lst:compare_one} demonstrates how the fallback mechanism provides an alternative in case of runtime errors in the auxiliary implementation. The \texttt{aux(x,y)} function converts variable \texttt{x} to a float, resulting in an operand type mismatch if \texttt{y} is a string. The main agent handles the type error by falling back to its secondary implementation. While the main implementation may still be vulnerable, this mechanism benefits small models that commonly encounter syntax and runtime errors. Despite some redundancy, a proper fallback can enhance code robustness.

\subsection{Decorator}

When the main agent trusts the auxiliary to accurately handle the core logic, it typically acts as a decorator, adding complementary features to enhance the output. The main agent could handle edge cases for improved robustness or reformat the outputs (e.g., sorting, rounding, and adjusting units). The necessity of such decorating logic depends on the auxiliary agent’s functionality and completeness.

\begin{listing}[H]
\caption{\texttt{compare\_one(a,b)}}
\label{lst:compare_one}
\begin{lstlisting}[language=Python]
def aux(x, y):
    # Convert x to a float
    if isinstance(x, str):
        x = float(x.replace(",", "."))

    # Vulnerable if y in different types
    if x < y:
        return y
    else:
        return x
    
def compare_one(a, b):
    try:
    	return aux(a, b)
	except Exception:
    	# Robust implementation
    	if type(a) == type(b):
        	if a > b:
            		return a
        	elif a < b:
            		return b
        	else:
            		return None
    	elif type(a) == float and type(b) == str:
        	b_float = float(b.replace(",", "."))
        	if a > b_float:
            		return a
        	elif a < b_float:
            		return b_float
        	else:
            		return None
    	elif type(a) == str and type(b) == float:
        		a_str = str(a.replace(",", "."))
        		a_float = float(a_str)
        	if a_float > b:
            		return a_str
        	elif a_float < b:
            		return b
        	else:
            		return None
    	else:
        	return None
\end{lstlisting}
\end{listing}

Listing~\ref{lst:palindrome} illustrates a decorator main function and how its utility varies depending on the completeness of the auxiliary implementation. In this problem, 2 agents generate codes together to check if a given string is a palindrome. When paired with \texttt{aux1} that only handles the recursion boundary condition of single-character strings, the empty string check of the main serves as a necessary edge case handler. However, when working with \texttt{aux2}, which already has a more comprehensive edge case consideration, this handle becomes redundant.

\subsubsection{Coordinator}

In large-scale software systems, it would be beneficial to have pipelines for repeated or data-parallel operations (e.g., batch processing, stream transformations). This corresponds to the coordinator cooperation scheme in our models, where the main agent divides the tasks into parts and assigns them to the auxiliary agent. 

A simple example involves the main agent acting as an iterator, using a loop (e.g., a \texttt{for} loop) to structure the problem. The auxiliary function generates partial solutions within each iteration. These partial results are then aggregated to form the final output. However, this cooperation scheme is unstable, as it depends heavily on the correct functionality of the auxiliary agent. When the auxiliary agent fails to complete its subtask, the entire solution breaks down.

\begin{listing}[H]
\caption{\texttt{is\_palindrome(text)}}
\label{lst:palindrome}
\begin{lstlisting}[language=Python]
def aux1(text):
    if len(text) == 1:
        return True
    return text[0] == text[-1] and aux1(text[1:-1])

def aux2(text):
    if len(text) <= 1:
        return True
    return text[0] == text[-1] and aux2(text[1:-1])

def is_palindrome(text):
    if not text:
        return True

    # Edge case handler
    return aux1(text)

    # Redundant decorator
    return aux2(text)
\end{lstlisting}
\end{listing}

\begin{listing}[H]
\caption{\texttt{flip\_case(string)}}
\label{lst:flip_case}
\begin{lstlisting}[language=Python]
def aux(string: str) -> str:
   result = ""
   for char in string:
       if char.islower():
           result += char.upper()
       elif char.isupper():
           result += char.lower()
       else:
           result += char
   return result

def flip_case(string: str):
   flipped = ""
   for char in string:
       flipped += aux(char)
   return flipped
\end{lstlisting}
\end{listing}

Listing~\ref{lst:flip_case} demonstrates a solution to flip the case of characters in a string. The auxiliary function flips the case of each character, while the main function calls this auxiliary function for each character and appends it to the result. This scheme can be extended to more complex scenarios, where subtasks are assigned in a hierarchical structure.

\subsubsection{Strategy Filter}

When handling complex problems, the main agent may need to implement logic based on multiple conditions. In such cases, the auxiliary agent can act as a filter for specific branches of logic, often appearing within conditional blocks (e.g., following an \texttt{if} statement). This scheme resembles the adaptive control flow in practice. In rule-based pipelines, an auxiliary agent evaluates preconditions (e.g., task types, system status, configurations) and directs workers to execute appropriate subroutines, thereby enhancing project modularity.

\begin{listing}[H]
\caption{\texttt{x\_or\_y(n,x,y)}}
\label{lst:nxy}
\begin{lstlisting}[language=Python]
def aux(n):
    if n < 2:
        return False
    if n == 2:
        return True
    if n % 2 == 0:
        return False
    for i in range(3, int(n**0.5) + 1, 2):
        if n % i == 0:
            return False
    return True

def x_or_y(n, x, y):
    # Check if n is prime
    if aux(n):
        return x
    else:
        return y
\end{lstlisting}
\end{listing}

Listing~\ref{lst:nxy} presents a solution for \texttt{x\_or\_y(n,x,y)} problem, which returns \texttt{x} if \texttt{n} is prime and \texttt{y} otherwise. The auxiliary function handles the primality checking, while the main function is responsible for returning results. The same pattern can also be found in the solutions of \texttt{prime\_fib(n)}, \texttt{factorize(n)}, and \texttt{largest\_prime\_factor(n)}.

\subsection{Learning Modes}

\begin{figure}[H]
    \centering
    \begin{subfigure}[b]{0.23\textwidth}
        \centering
       \includegraphics[width=\textwidth]{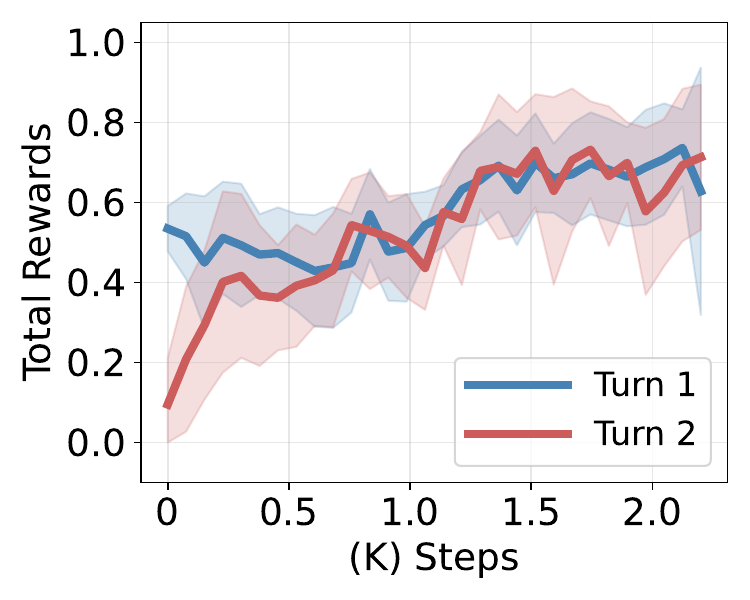}
        \caption{Self-Improvement}
        \label{subfig:ls_1}
    \end{subfigure}
    \begin{subfigure}[b]{0.23\textwidth}
        \centering
        \includegraphics[width=\textwidth]{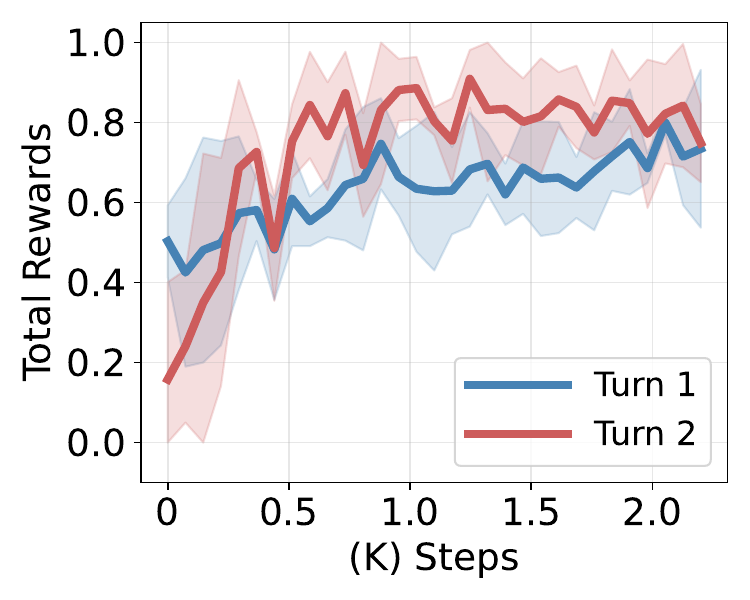}
        \caption{Guided-Learning}
        \label{subfig:ls_2}
    \end{subfigure}
    \caption{Learning modes in 2-turn MAGRPO training.}
    \label{fig:learning_scheme}
\end{figure}

Figure~\ref{fig:learning_scheme} shows the reward curves of total returns in a 2-turn MAGRPO training with 2 different learning modes. 

Figure~\ref{subfig:ls_1} demonstrates a self-improvement mode, where agents interact with the tasks themselves and do not have access to the external feedback. At each turn $t$, the prompts are just prompts and responses from the previous turns and a revision instruction. Although both curves improve as agents gradually develop cooperative behaviors, the performance of the second turn is still consistently similar to the first turn, suggesting that learning is primarily driven by direct task interaction.

Figure~\ref{subfig:ls_2} illustrates the reward curves of guided-learning mode, where LLMs leverage external feedback to improve performance, e.g., feedback from expert agents, diagnosis from static checkers like \textit{AST}, and sandbox tests. When employing \textit{Claude-Sonnet-4} as an external model to provide more concrete suggestions or feeding error messages from sandbox tests, the performance of the second turn (red) exceeds first turn (blue), and both outperform those in the self-improvement. This indicates that appropriate guidance helps agents to refine the response in an efficient way. Due to the computational constraints, most models used in our setup have around 3B parameters and may struggle to interpret more abstract suggestions. We hypothesize that larger models with higher reasoning capabilities could benefit from more implicit guidance.

\section{Experimental Configurations}

This section outlines the hyperparameter settings and reward specifications used in our experiments.

\subsection{Hyperparameters}

For writing collaboration, we set the temperature to $0.8$ and apply nucleus sampling with a threshold of $0.95$ to encourage diverse generation. Policy deviation is regularized using a beta value of $0.02$. The policy is optimized using a learning rate of $5 \times 10^{-6}$, and training is conducted for 1,500 steps.

For coding collaboration, the single-turn MAGRPO training uses a temperature of $0.7$ and nucleus sampling with a threshold of $0.9$. The learning rate is set to $1 \times 10^{-6}$, with 1,500 training steps. In the multi-turn setting, the discount factor is set to $1.0$, and the learning rate is $5 \times 10^{-6}$, with 2,200 training steps.

\subsection{Reward Specifications}

Rewards are computed as a weighted sum of multiple metric-based components, following a hierarchical reward modeling scheme to prioritize fundamental objectives.

\subsubsection{TLDR Summarization}

\begin{itemize}
	\item \textbf{Structural Wellness}: The structural wellness is assessed by the ratios of paragraph length and unique words. For the completion length, an appropriate ratio within 1.6-3.2$\times$ receives the full rewards; ratios within the range of 1.1-5.0$\times$ receive proportional rewards; while ratios outside receive no rewards and early termination of evaluation. For the ratio of unique words, we exclude the common stopwords. A ratio of 2.0$\times$ or higher receives the maximum rewards; ratios between 1.3-2.0$\times$ receive proportional rewards; ratios below 2.0$\times$ result in no rewards and evaluation termination.
	\item \textbf{Style Consistency}: The style consistency is measured through Jaccard similarity of vocabulary between the completions (excluding stopwords). The Jaccard similarity scores are capped at $0.03$ and normalized as rewards. We use a cap here to balance the needs of maintaining lexical consistency and vocabulary expansion in elaborated summarization.
	\item \textbf{Logical Coherence}: The logical coherence is evaluated through the presence and diversity of transition words in the completions. We check transition words across 12 categories, e.g., examples, explanation, contrast, etc. Additional rewards are given for using transition words in more categories, where $r=\min(0.6\log(n+1), 1)$, and $n$ is the number of transition categories.
\end{itemize}

\subsubsection{arXiv Expansion}

\begin{itemize}
	\item \textbf{Structural Wellness}: This metric evaluates the relative length and lexical diversity between the second and first completions. A length ratio within the optimal range of 1.0–1.3$\times$ yields the maximum rewards, while ratios within the acceptable bounds of 0.8–1.5$\times$ receive proportionally scaled rewards. Ratios outside this range result in zero reward and early termination. Similarly, a unique word ratio within 0.7–1.3$\times$ receives the full rewards, ratios within 0.5–1.7$\times$ are rewarded proportionally, and values outside this range lead to zero reward and evaluation termination.
	\item \textbf{Style Consistency}: Style consistency is quantified using Jaccard similarity between the 2 completions. The similarity score is capped at 0.23 and normalized as rewards.
	\item \textbf{Logical Coherence}: Logical coherence is assessed based on the presence of transition words across 12 categories. Additional rewards are given for using transition words in more categories, where $r=\min(0.4\log(n+1), 1)$, and $n$ is the number of transition categories.
\end{itemize}

\subsubsection{Coding Collaboration}

\begin{itemize}
\item \textbf{Structural Integrity}: This metric verifies the correct implementation of both the auxiliary and main functions. To receive the base reward, the corresponding functions in the agents’ completions must be properly defined and include return statements. Failure to define the main function results in evaluation termination.
\item \textbf{Syntactical Correctness}: This metric assesses the syntactic validity of the concatenated code, which includes the libraries provided in the dataset, the auxiliary function defined by the helper agent, and the function defined by the main agent. Syntactic correctness is verified via static analysis, i.e., Abstract Syntax Tree (AST). The presence of syntax errors leads to the termination of the evaluation to avoid runtime failures.
\item \textbf{Test Pass Rate}: This metric measures the percentage of unit tests passed during execution, with each test subject to a 10-second timeout. Rewards are assigned proportionally based on the number of successful assertions. If no tests pass, the evaluation is terminated.
\item \textbf{Cooperation Quality}: A base bonus is applied if the main function calls the auxiliary. Additional rewards are given when the main function implements substantive logic beyond simply wrapping the auxiliary.
\end{itemize}

\section{Prompt Design}

\subsection{Writing Collaboration}

\subsubsection{TLDR} In the TLDR summarization, the \texttt{prompt} field of the dataset is processed for each agent by using the following instructions.

\begin{lstlisting}[numbers=none, escapeinside={(*}{*)}]

(*\textbf{Summary Agent}*) 
Create a concise summary response to this post.
Query: {prompt}
Instructions: Provide a brief and focused summary in a few sentences

(*\textbf{Elaboration Agent}*) 
Create a detailed summary response to this post.
Query: {prompt}
Instructions: You should use transition words to improve flow
\end{lstlisting}

\subsubsection{arXiv} In the arXiv paragraph expansion, we use the \texttt{abstract} field of the dataset and process it as follows. 

\begin{lstlisting}[numbers=none, escapeinside={(*}{*)}]

(*\textbf{Background Agent}*) 
Based on the following scientific abstract, expand the content for the introduction section.
Abstract: {abstract}
Instructions:
- There is another agent that will provide the method and implications
- You just need to focus on the background and motivation
- Avoid repeating methodology and implications content

(*\textbf{Method Agent}*) 
Based on the following scientific abstract, expand the content for the introduction section.
Abstract: {abstract}
Instructions:
- There is another agent that will provide the background and motivation
- You just need to focus on the method and implications
- Avoid repeating background and motivation content
\end{lstlisting}

\subsection{Coding Collaboration}

For HE and CHE, we extract the \texttt{entry\_point}, \texttt{params} from the \texttt{prompt} field and instruct agents as follows.

\begin{lstlisting}[numbers=none, escapeinside={(*}{*)}]

(*\textbf{Auxiliary Agent}*) 
Create a helper function for this coding problem.
Problem: {prompt}
Instructions:
- Output ONLY the function code, no explanations or examples
- Do NOT include markdown code blocks (```python)
- Do NOT include any text before or after the function
- Do NOT include test cases or example usage
- Create a helper function named 'aux' that can assist the main function
- The function should return useful data for solving the problem

Your output should follow this format:
def aux(...):
    # your code here
    return result
    
(*\textbf{Main Agent}*) 
Solve this coding problem by implementing the required function.
Problem: {prompt}
You have access to a helper function: aux(...)

Instructions:
- Output ONLY the function code, no explanations or examples
- Do NOT include markdown code blocks (```python)  
- Do NOT include any text before or after the function
- Do NOT include test cases or example usage
- Do NOT redefine the aux() function
- Implement ONLY the '{entry_point}' function as specified
- You can call aux() to assign a value to a variable within your function if helpful

Your output should follow this format:
def {entry_point}({params}):\n # your function code here\nreturn result\n
\end{lstlisting}

To improve the generated code, these prompts are used to construct second-turn observations for the MAS with suggestions from \textit{Claude-Sonnet-4}.

\begin{lstlisting}[numbers=none, escapeinside={(*}{*)}]

(*\textbf{External Agent}*)
You are an advisor helping 2 agents (an auxiliary agent and a main agent) solve the following problem. The auxiliary agent provides a helper function (aux), while the main agent defines the task-specific logic.
Problem: {prompt}
Example tests: {test}
Show your feedback and edits for the following code: {combined_code}

Instructions: 
- If you identify a missing element, such as an undefined aux or missing entry point (main function), you should write one for it. 
- If both are not missing, point out and make changes to any critical syntax or logic errors that would prevent the code from passing the given unit tests.
- You should focus only on clear errors on the given unit tests, be conservative and lenient, ignoring issues like redundancy, inefficiency, lack of edge case handling, or type annotations. 
- Return "Perfect! No edits needed!" if the logic is sound.

Your response MUST contain the JSON format specified below. Always include both 'aux' and 'main' fields, even if no edits are needed.
{ "aux": {{...}}, "main": {{...}}}
\end{lstlisting}

\section{Baseline Methods}

We adopt a single-agent method and 3 representative multi-agent conversation methods as baselines.
\begin{figure}[H]
    \centering
    \includegraphics[width=0.4\textwidth]{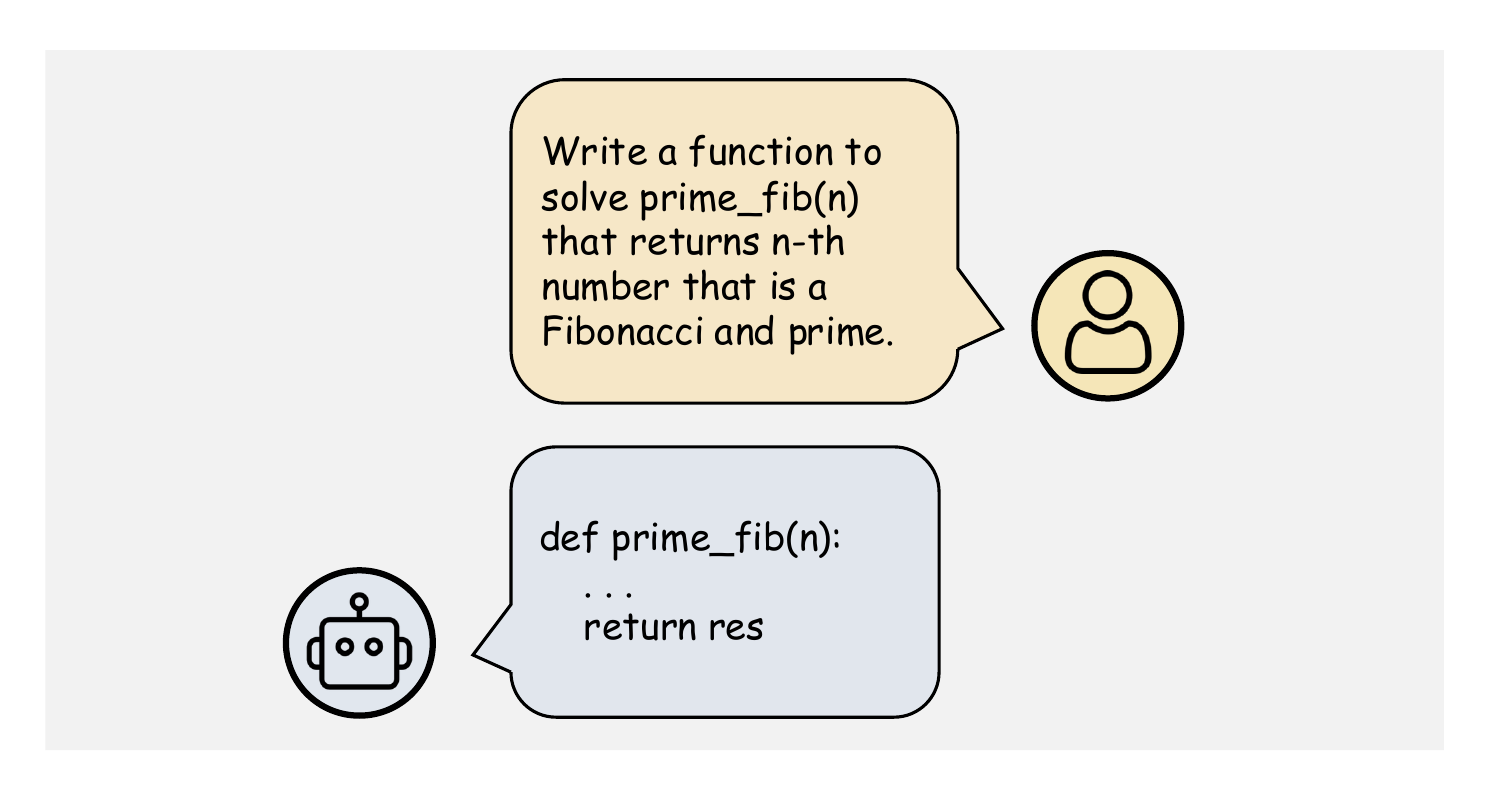}
    \caption{Single-agent code generation.}
    \label{fig:baseline_single}
\end{figure}
Figure~\ref{fig:baseline_single} illustrates the code generation process using a single LLM agent. In this setting, the user gives a coding question along with specific instructions. The agent responds by generating a Python function snippet to solve it.
\begin{figure}[H]
    \centering
    \includegraphics[width=0.4\textwidth]{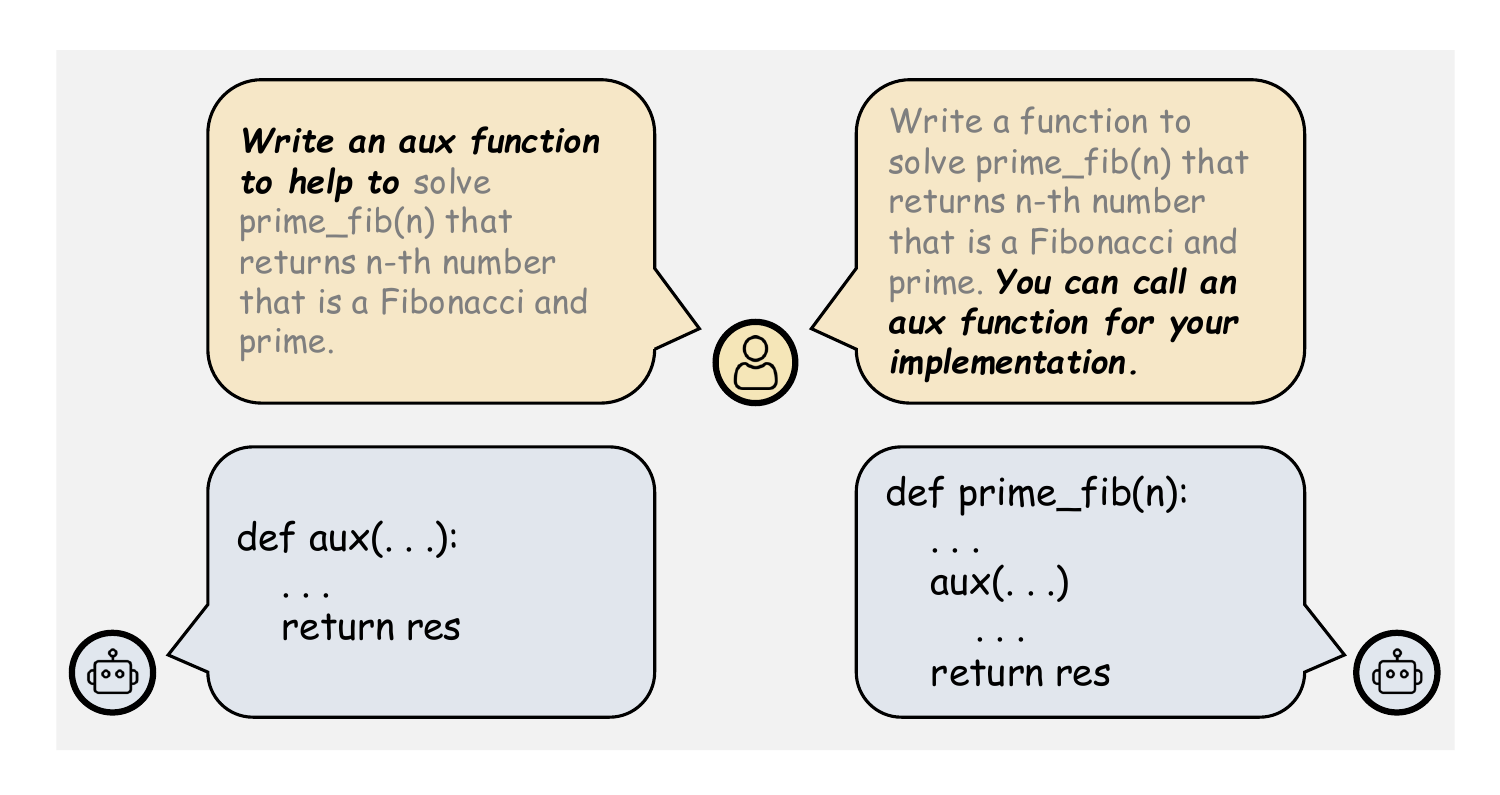}
    \caption{Coding collaboration via naive concatenation.}
    \label{fig:baseline_concat}
\end{figure}
Naive concatenation represents the simplest form of cooperation, where 2 agents generate code synchronously, as illustrated in Figure~\ref{fig:baseline_concat}. The first agent is provided with the coding question and informed of its role as a helper. The second agent is given the same question, along with its role as the main generator and the fact that an auxiliary agent exists. However, the main agent lacks information about the specific functionality of the auxiliary agent. Their outputs are directly concatenated to form the response. This method is intended to improve inference efficiency by enabling a simple division of the problem into separate parts.
\begin{figure}[H]
    \centering
    \includegraphics[width=0.4\textwidth]{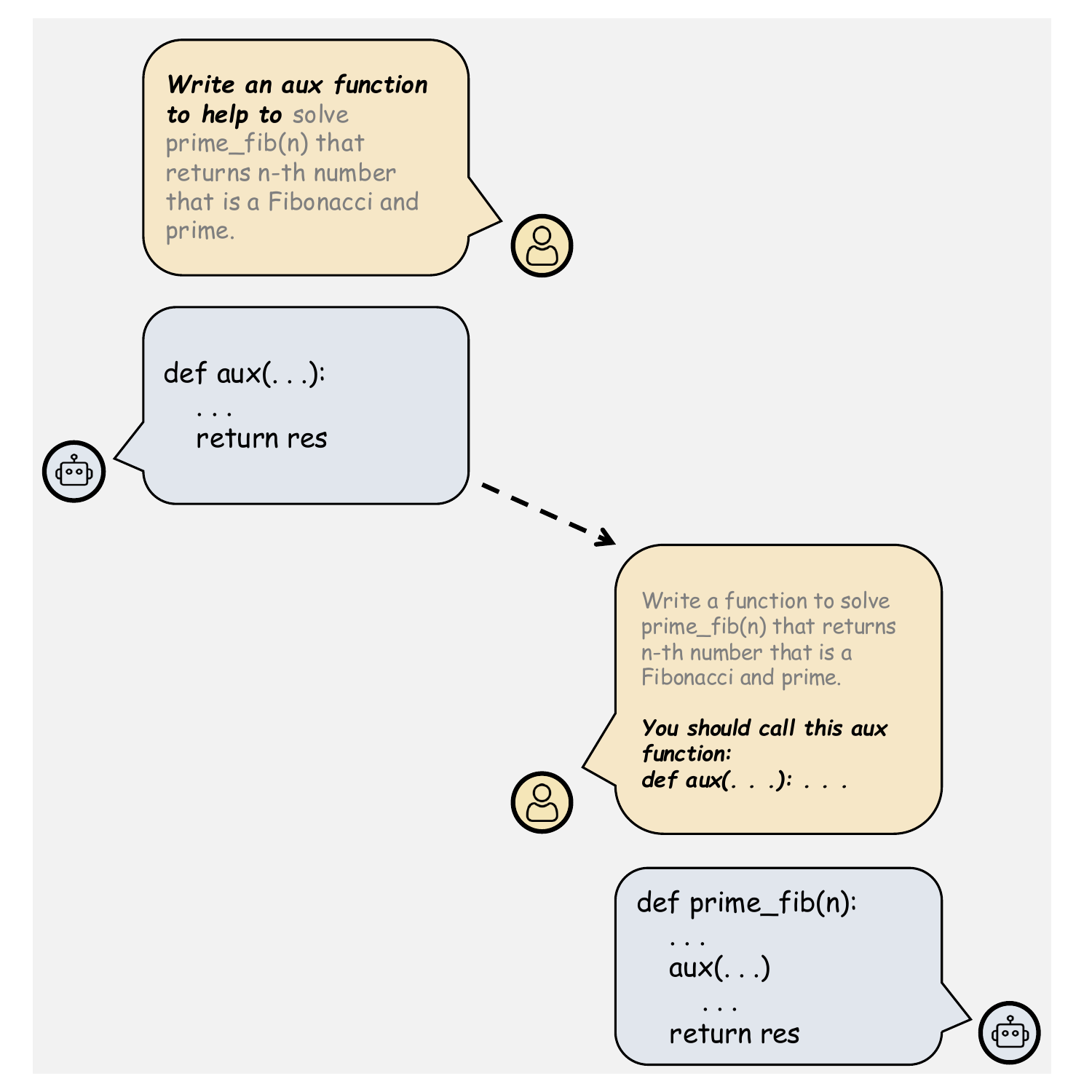}
    \caption{Coding collaboration via sequential pipeline.}
    \label{fig:baseline_sequential}
\end{figure}
Figure~\ref{fig:baseline_sequential} presents the form of pipeline cooperation, where agents respond in sequence. The first agent is given the coding question along with the role of a helper. Its response is then passed to the main agent as a reference. The main agent generates its answer by incorporating the helper’s response. This method enables one-way communication, allowing the main agent to respond by coordinating with the helper. However, this comes at the cost of slower inference speed due to the sequential nature of the interaction.
\begin{figure}[H]
    \centering
    \includegraphics[width=0.4\textwidth]{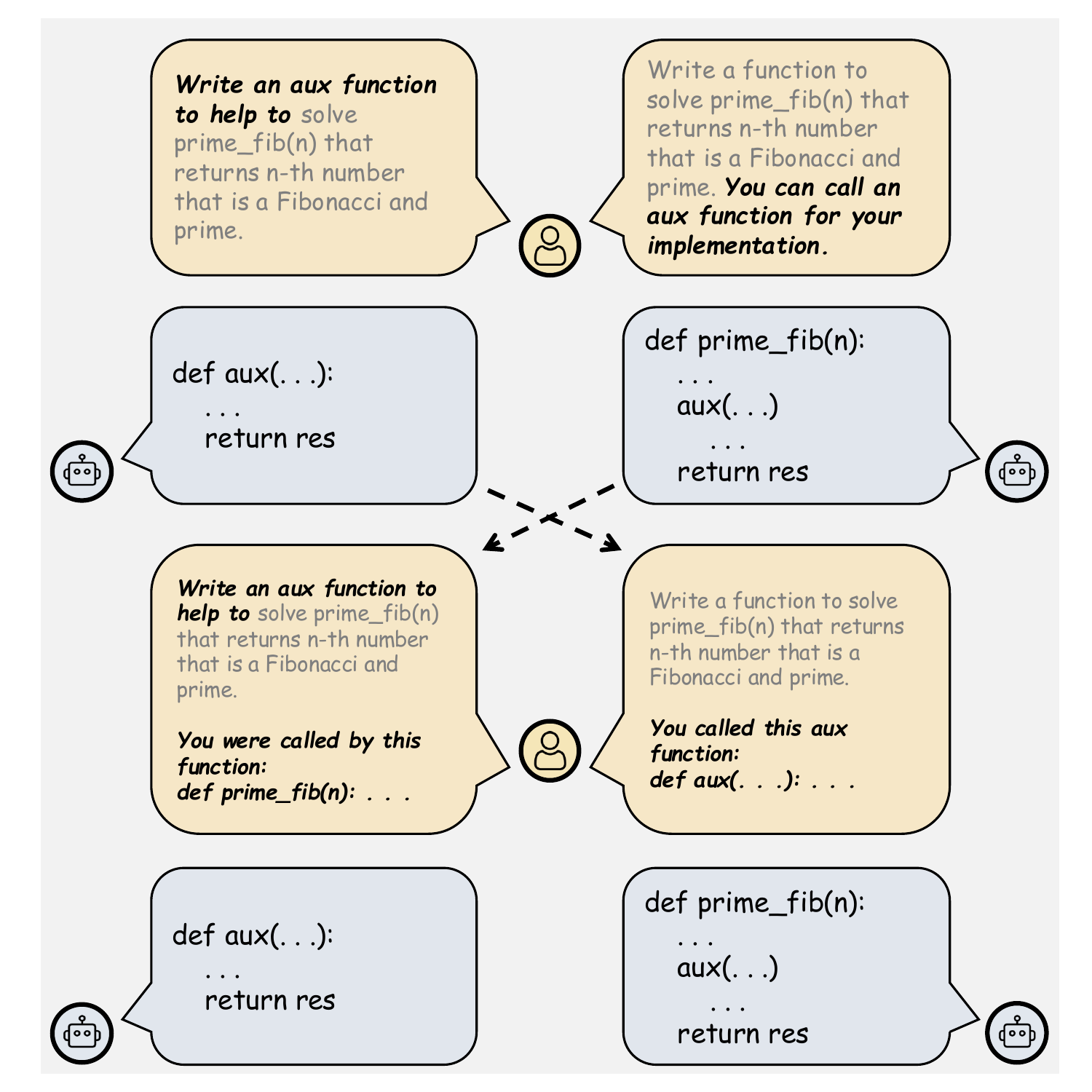}
    \caption{Coding collaboration with one-round discussion.}
    \label{fig:baseline_discussion}
\end{figure}
Discussion or debate frameworks (Figure~\ref{fig:baseline_discussion}) aim to improve response quality by enabling agents to access each other’s previous outputs \cite{dudebate, liangdebate}. In the first turn, the helper and main agents generate responses in the same manner as the naive concatenation approach. These initial responses are then shared with each other as references for the following turns, forming a discussion. Although this setup introduces more interaction, it does not guarantee improved response quality. With a limited number of rounds, the final output may not converge to a coherent solution. Even with additional rounds, convergence remains uncertain. This approach can even be less efficient than the sequential pipeline, particularly in distributed systems where communication latency is high or unreliable.

\begin{figure}[H]
    \centering
    \includegraphics[width=0.4\textwidth]{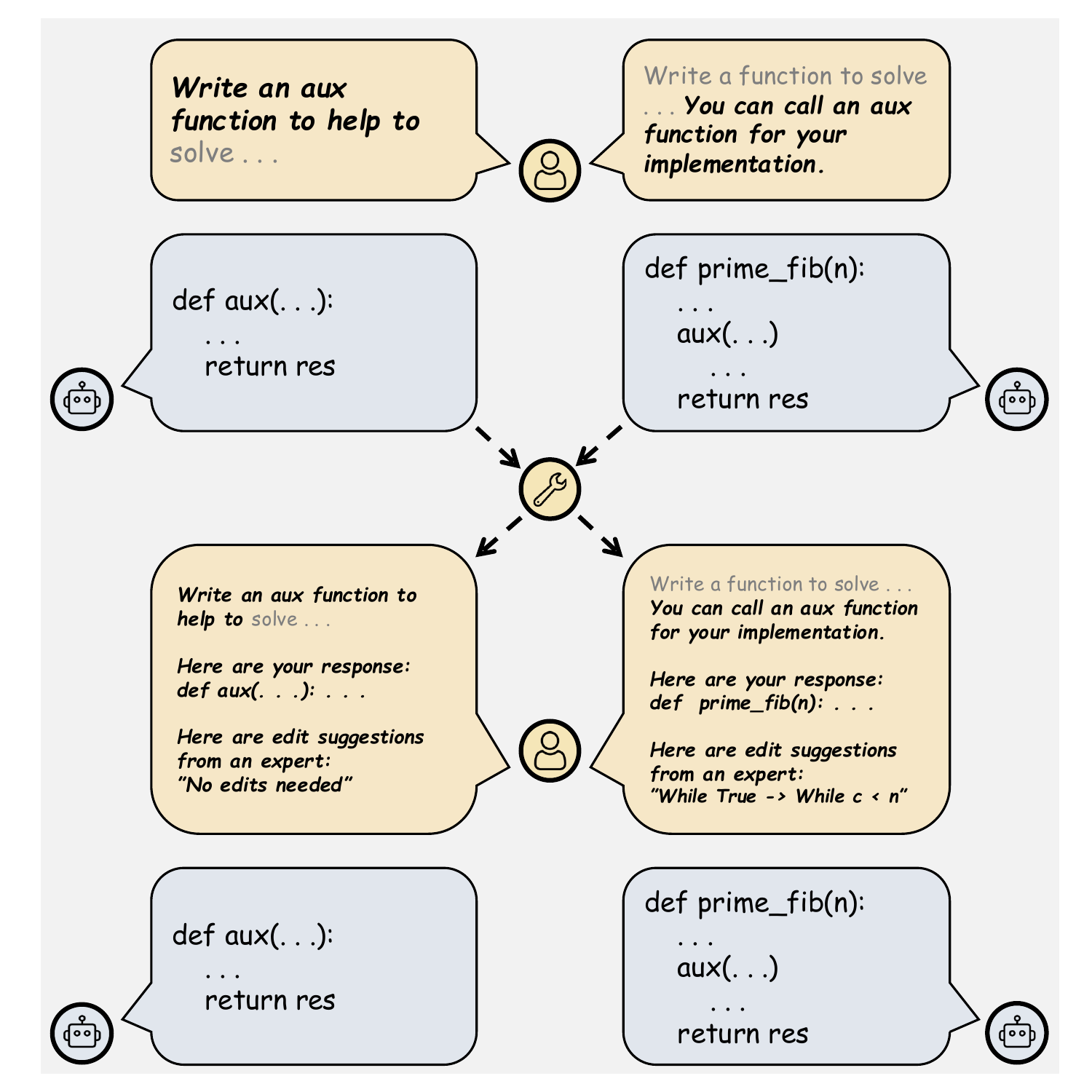}
    \caption{Coding collaboration in our method.}
    \label{fig:ours}
\end{figure}
The interaction process between 2 agents trained with MAGRPO is illustrated in Figure~\ref{fig:ours}. In the single-turn setting, we use the same prompts as in the naive concatenation baseline. In the multi-turn setting, after the helper and main agents generate their initial responses, these outputs are reviewed by an external agent. In this work, we employ \textit{Claude-Sonnet-4} as an external to provide edit suggestions. For each agent, the suggestions, as well as the prior information and their previous response, are incorporated into the prompt for the subsequent round.

Note that the baselines above can also be fine-tuned by MARL. However, the interactions among agents in these settings are not strictly cooperative, which may lead to instability during training. To address this, techniques such as role-based rewards \cite{spiral}, partial MAS training \cite{trainaleader}, and freezing selected agents \cite{maft} can be employed to ensure stable optimization.

\section{Broader Impacts}

Prompt-based coordination is often brittle \cite{debatefail}, as agents may fail to follow instructions they were not explicitly trained to interpret. Our method builds on a solid theoretical foundation in cooperative MARL, explicitly optimizing agents for joint optimality. Our work also opens opportunities to enhance existing test-time multi-agent interaction methods by integrating MARL techniques \cite{dudebate, duverifier, autogen}, particularly in settings that involve task decomposition and iterative feedback integration.

This work also explores a new perspective on accelerating LLM inference through cooperative MARL. While mainstream acceleration techniques (e.g., knowledge distillation, pruning, and quantization) improve efficiency at the cost of information loss \cite{KDinformationloss, pruninginfoloss}, our approach suggests decentralized coordination among specialized agents, thereby alleviating the burden of long-context memory and joint decision-making on a single model. Each agent can focus on a specific subtask, enabling more modular and robust reasoning.

\section{Limitations and Future Works}

Nevertheless, this study is subject to several limitations:

\begin{enumerate}
    \item First, we focus on homogeneous agents for simplicity, assuming they perform similar tasks despite being assigned different roles, e.g., both the auxiliary agent and main agent are generating Python functions. Future research could explore LLM collaboration among heterogeneous agents with diverse capabilities and functionalities.
    \item Due to computational constraints, we train LLMs with MAGRPO on limited datasets using relatively small-scale language models in a short horizon. When LLM-based coding agents are deployed in larger-scale projects involving multiple files and modules, more diverse and complex cooperation schemes are likely to emerge, which would further demonstrate the potential of decentralized coordination in MAS.
    \item The simplicity of our reward model inevitably leads to narrow reward signals and potential reward hacking. As suggested by many research studies and industrial practice \cite{prm, multi-aspect, anthropic2023collective}, designing more expressive and fine-grained reward models (e.g., multi-aspect rewards, process-supervised rewards) is essential for better aligning agent cooperation with human preferences.
\end{enumerate}

\section{Compute Resources}

We use NVIDIA H200 and H100 GPUs for LLM fine-tuning, and a standalone NVIDIA GeForce RTX 5090 workstation for inference. Here are the specifications of the resources used for experiments.

\begin{lstlisting}[numbers=none, escapeinside={(*}{*)}]

(*\textbf{Training Devices}*) 
Type: GPU Cluster
CPU:  Intel Xeon Platinum 8558
GPU:  1x NVIDIA H200

(*\textbf{Inference Device}*) 
Type: Standalone Workstation
CPU:  AMD Ryzen 9 9950X
GPU:  2x NVIDIA GeForce RTX 5090
\end{lstlisting}
\vspace{3mm}

The training process requires approximately 5 hours to complete 700 steps and 8 hours to complete 1000 steps. The inference takes approximately 0.5 hours for each 15-generation evaluation. Noted that training duration may vary considerably due to the node condition and the stochasticity of LLM outputs.


\section{Codes and Datasets}

\paragraph{Codes} We developed an open-source library, CoMLRL, for training multiple LLMs to collaborate using MARL. CoMLRL provides configurable implementations of various MARL algorithms for LLM collaboration, including MAGRPO. The experiments of this paper serve as parts of CoMLRL's environments and benchmarks.

\begin{itemize}
    \item \textbf{\texttt{OpenMLRL/CoMLRL}:} 
    
    \url{https://github.com/OpenMLRL/CoMLRL}
    
    \item \textbf{\texttt{OpenMLRL/LLM\_Collab\_Writing}:}
    
    \url{https://github.com/OpenMLRL/LLM_Collab_Writing}
    
    \item \textbf{\texttt{OpenMLRL/LLM\_Code\_Generation}:} 
    
    \url{https://github.com/OpenMLRL/LLM_Collab_Code_Generation}
    
    \item \textbf{\texttt{OpenMLRL/LLM\_Code\_Completion}:}
    
    \url{https://github.com/OpenMLRL/LLM_Collab_Code_Completion}
    
\end{itemize}

We hope that this library and the associated experiment repositories will lay a foundation and provide convenient tools for future research and development in LLM collaborations.
 
\paragraph{Datasets} The datasets used in our experiments are all open-sourced and available online,

\begin{itemize}
    \item \textbf{\texttt{trl-lib/TLDR}:} 
    
    \url{https://huggingface.co/datasets/trl-lib/tldr}
    
    \item \textbf{\texttt{mattbierbaum/arXiv}:} 
    
    \url{https://github.com/mattbierbaum/arxiv-public-datasets}
    
    \item \textbf{\texttt{OpenAI/humanevalHumanEval}:}
    
    \url{https://huggingface.co/datasets/openai/openai_humaneval}
    
    \item \textbf{\texttt{OpenMLRL/CoopHumanEval}:}
    
    \url{https://huggingface.co/datasets/OpenMLRL/CoopHumanEval}
\end{itemize}

\end{document}